\documentclass{article}

\usepackage[numbers,square]{natbib}
\usepackage[final,main]{neurips_2026}

\usepackage[utf8]{inputenc}
\usepackage[T1]{fontenc}
\usepackage[colorlinks,urlcolor=blue,linkcolor=blue,citecolor=blue]{hyperref}
\usepackage{url}
\usepackage{graphicx}
\usepackage{booktabs}
\usepackage{amsmath,amssymb}
\usepackage{amsfonts}
\usepackage{nicefrac}
\usepackage{microtype}
\usepackage[table]{xcolor}
\usepackage{multicol}
\usepackage{multirow}
\usepackage{wrapfig}
\usepackage{subcaption}
\usepackage[capitalize,nameinlink]{cleveref}
\usepackage{tabularx}
\usepackage{bbding}

\definecolor{slateL}{HTML}{F3F5F7} 
\definecolor{slateM}{HTML}{E5EAF0} 
\definecolor{slateH}{HTML}{D0D7E3} 

\newcommand{\ie}{i.e.}

\title{
Resolving Representation Ambiguity in Feedforward Novel View Synthesis Transformer \\ via Semantic-Spatial Decoupling} 

\author{%
  Yihang Wu\textsuperscript{1,2}\thanks{This work is supported by the National Natural Science Foundation of China (Grant No. 62427819), the Science and Technology Commission of Shanghai Municipality (No. 24511103100), and the New Cornerstone Science Foundation through the XPLORER PRIZE.} \quad
  Yihang Sun\textsuperscript{3*} \quad
  Shaofeng Zhang\textsuperscript{4} \And
  Zuxuan Wu\textsuperscript{1,2} \quad
  Junchi Yan\textsuperscript{3}\quad
  Xiaosong Jia\textsuperscript{1,2}\textsuperscript{\Envelope} \quad Yu-Gang Jiang\textsuperscript{1,2}\\\\
  $^{1}$Institute of Trustworthy Embodied Artificial Intelligence (TEAI), Fudan University \\
  $^{2}$Shanghai Key Laboratory of Multimodal Embodied AI\\
  $^{3}$Sch. of Artificial Intelligence \& Sch. of Computer Science, Shanghai Jiao Tong University\\
  $^{4}$University of Science and Technology of China\\
  * Equal Contributions \quad\quad
  \textsuperscript{\Envelope} Corresponding Authors
}

\begin{document}

\maketitle

\begin{abstract}
  Transformer-based models have advanced feedforward novel view synthesis (NVS). Current architectures such as GS-LRM and LVSM mix semantic information (e.g., RGB) and spatial information (e.g., Pl\"ucker rays) into a shared feature space. Since Pl\"ucker rays naturally carry lattice-like spatial structure, these designs can make the spatial bias interfere with appearance representation and degrade rendering fidelity. To this end, we propose to \textbf{decouple the representation of feedforward NVS transformers into separate semantic and spatial tokens}. The decoupled design keeps semantic and spatial information explicit in their branches while preserving cross-branch interaction through shared attention routing. Built on this design, we introduce optional categorized supervision and bidirectional modulation: the former provides branch-specific training signals, while the latter improves interaction between the two branches. Notably, the base decoupled design introduces virtually zero additional inference latency due to its architectural design. The proposed designs achieve consistent improvements, demonstrating effectiveness across decoder-only and encoder-decoder feedforward NVS models.

\end{abstract}

\begin{center}
  \small Project page: \url{https://hangzay.github.io/ssd_lvsm/}
\end{center}

\begin{figure}[!t]
    \centering
    \includegraphics[width=1.0\linewidth]
    {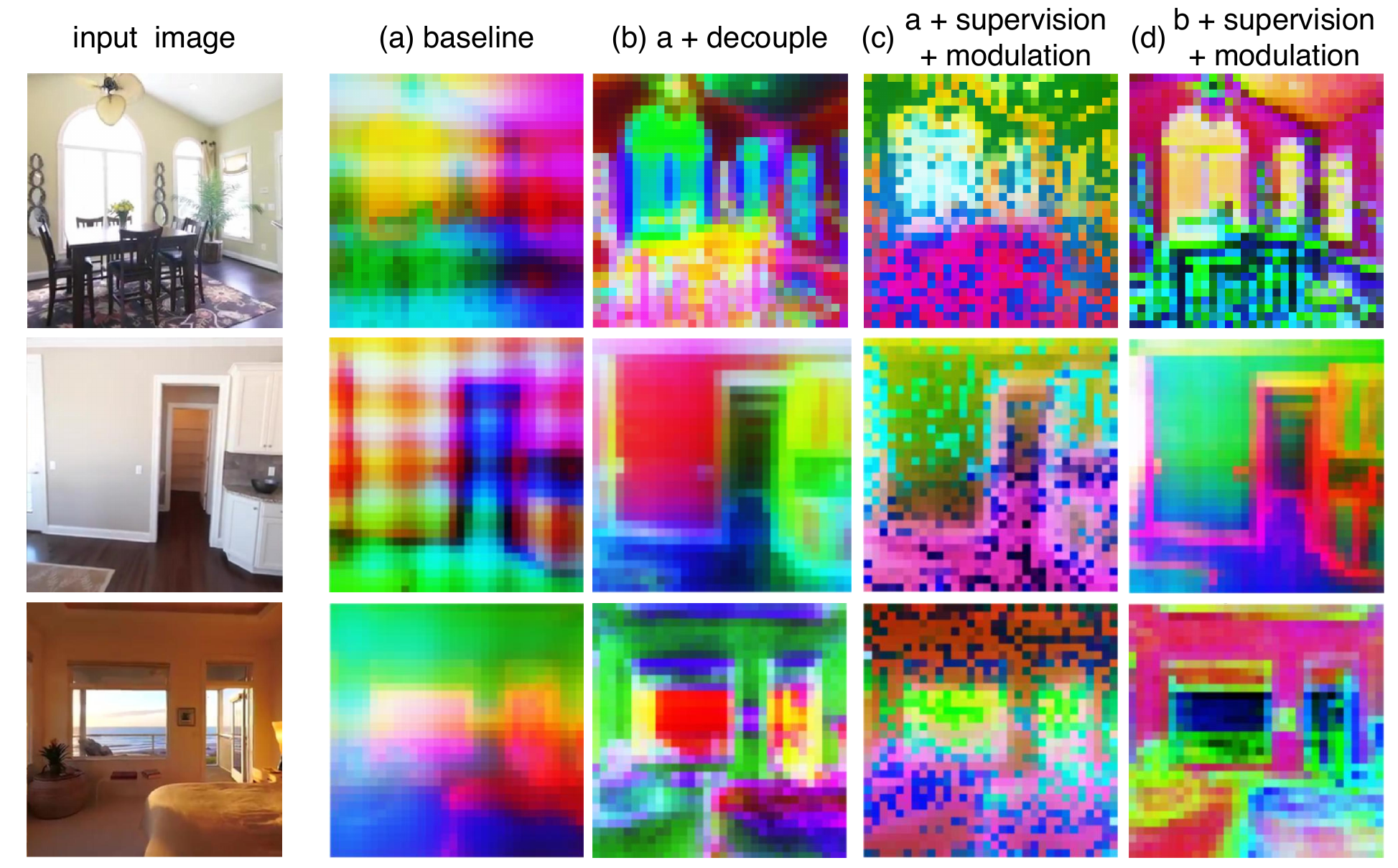}
    \vspace{-5mm}
    \caption{\textbf{Feature Map Comparison across Model Variants.}
    We visualize middle-layer LVSM features: (a) the LVSM baseline;
    (b) Decouple-only, with separate semantic and spatial branches and no
    categorized supervision or modulation; (c,d) supervision and modulation
    added to (a,b), respectively. Decoupling produces more structured
    representations and enables the auxiliary designs.}

    \vspace{-5mm}
    \label{fig:compare}
\end{figure}

\section{Introduction}
\label{sec:intro}

Novel View Synthesis (NVS) has undergone a fundamental paradigm shift, moving from per-scene optimization~\cite{mildenhall2020nerf,kerbl20233d} to generalizable feedforward inference powered by Large Reconstruction Models (LRMs)~\cite{hong2023lrm}.
Transformer-based architectures like GS-LRM~\cite{zhang2024gslrm} and LVSM~\cite{jin2024lvsm} typically adopt a unified token approach, where semantic features (RGB patches) and spatial features (Pl\"ucker rays) are concatenated and projected into a shared latent space.
These inputs serve different roles: RGB provides appearance evidence, while camera rays describe the viewing geometry associated with each patch.
However, since Pl\"ucker-ray features carry lattice-like spatial structure, jointly propagating both types of information through shared feature transformations can make spatial bias interfere with appearance representation.
Additionally, we observe grid-like distortions in intermediate features (Fig.~\ref{fig:compare}), resembling artifacts previously reported in vision Transformers~\cite{shi2024denoising}.
This motivates us to examine how appearance and spatial conditioning interact within the token representation.

\begin{figure}[!t]
    \centering
    \includegraphics[width=1.0\linewidth]
    {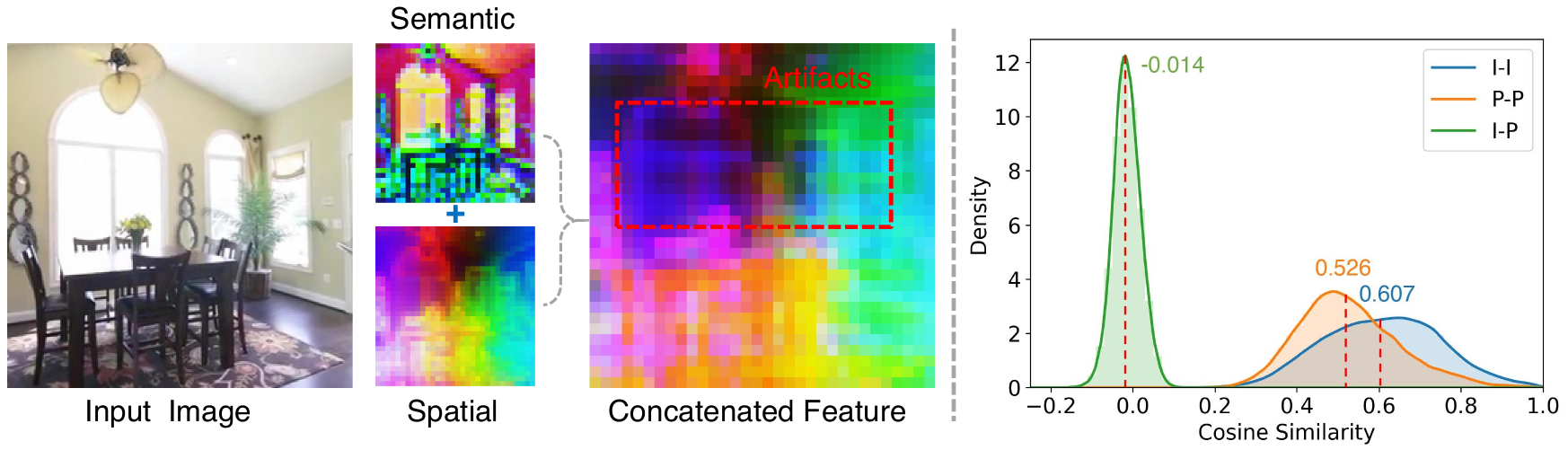}
    \vspace{-6mm}
    \caption{\textbf{Feature Coupling and Cosine Similarity.} Left: Semantic and spatial features are structured separately, while their concatenation exposes grid-like Pl\"ucker-related artifacts. Right: We sample vector pairs to estimate cosine similarity distributions for I--I (semantic), P--P (spatial), and I--P (cross-branch). Dashed lines indicate the mean of each group.}

    \vspace{-5mm}
    \label{fig:cos}
\end{figure}

\begin{figure}[t]
    \centering
    \includegraphics[width=1.0\linewidth]{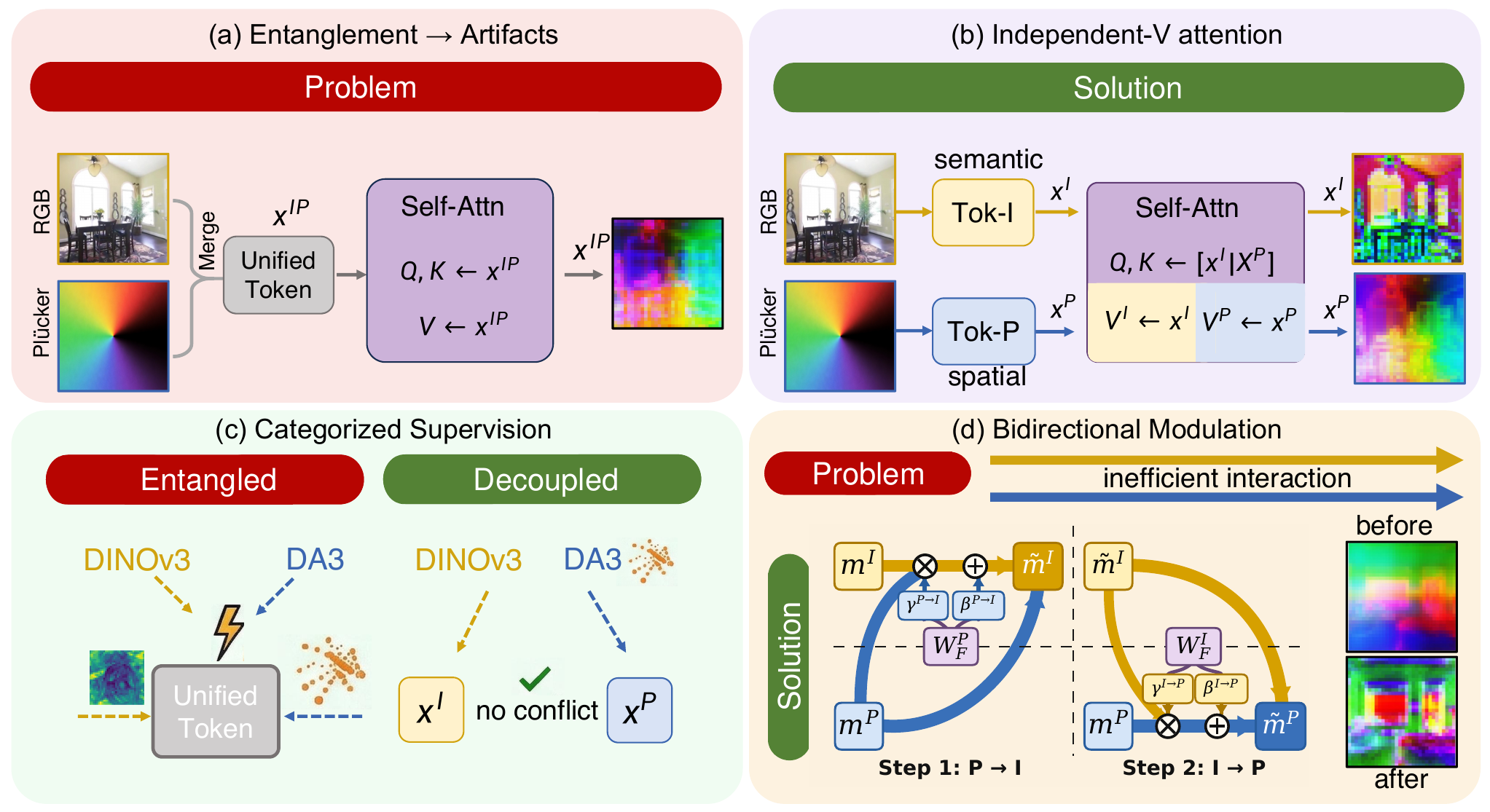}
    \vspace{-6mm}
    \caption{\textbf{Overview.} In this work, we identify that mixing RGB and Pl\"ucker-ray information in a shared feature space can make spatial bias interfere with appearance representation. We decouple the two information streams while preserving cross-branch interaction to enhance rendering.}
    \vspace{-8mm}
    \label{fig:overview}
\end{figure}

Inspired by register tokens~\cite{darcet2023vision} in ViTs, we first introduce registers to provide extra storage slots, but this control does not resolve the issue. The persistence of grid-like patterns when RGB inputs are removed further points to the role of spatial conditioning. In this work, we propose to \textbf{decouple semantic and spatial tokens}, redesigning the Transformer representation space to preserve semantic and spatial feature independence while facilitating cross-stream context exchange. As shown in Fig.~\ref{fig:cos}, the two branches exhibit different feature patterns. The core architectural change is full-pipeline decoupled Transformer by \textbf{Independent-V Attention mechanism}: shared query-key routing coordinates the branches, while separate value projections, layer normalization, and feed-forward networks keep their feature updates branch-specific. Based on this decoupled representation, we further introduce optional categorized supervision and bidirectional modulation as auxiliary designs (Fig.~\ref{fig:overview}). The supervision terms provide branch-specific training signals and are removed at inference, while modulation improves controlled cross-branch conditioning.

Notably, the base decoupled design incurs virtually zero additional inference latency while improving rendering quality over the entangled baseline. In our controlled RealEstate10K setting, optional supervision and modulation further improve performance, reaching a 1.1\,dB PSNR gain with an 8\% inference-time increase. Extensive experiments across decoder-only and encoder--decoder architectures, larger backbones, and scene- and object-level datasets validate the effectiveness of the proposed approach.
The benefit also persists with raw raymaps, indicating that semantic--spatial decoupling is not specific to the Pl\"ucker parameterization.

\section{Related Work}

\textbf{Feedforward Novel View Synthesis.}
NVS has progressed from image-based rendering~\cite{debevec1996modeling} and per-scene neural optimization~\cite{mildenhall2020nerf, kerbl20233d} to feed-forward approaches.
Early methods~\cite{yu2021pixelnerf, wang2021ibrnet, chen2021mvsnerf} conditioned NeRF networks on pixel-aligned features but retained explicit 3D inductive biases, whereas transformer-based LRMs~\cite{hong2023lrm, zhang2024gslrm, ziwen2024longlrm} learn scalable reconstruction priors from data.
Models without explicit 3D representations~\cite{jin2024lvsm, jiang2025rayzer, jia2026efficientlvsm, liu2025kaleido} further simplify NVS into token-level image prediction.
Generation-based approaches such as Seva~\cite{zhou2025seva}, GEN3C~\cite{gen3c}, and CAT3D~\cite{cat3d} synthesize camera-controlled views, while Lyra and Lyra~2.0~\cite{lyra,lyra2} use video priors to construct 3D scenes. LGM~\cite{lgm}, TRELLIS~\cite{trellis}, and Hunyuan3D~2.1~\cite{hunyuan3d21} focus on 3D content or asset creation; these settings differ in inputs and outputs from posed-image reconstruction.
Despite these advances, recent feedforward NVS transformers commonly encode semantic appearance and spatial geometry within unified tokens.
Our work addresses the resulting intra-token representation ambiguity, complementary to concurrent efforts on improved camera conditioning~\cite{wu2026rays}.

\textbf{Token Dynamics and Feature Artifacts in Vision Transformers.}
ViTs~\cite{dosovitskiy2020vit} exhibit representational artifacts such as grid-like patterns linked to positional encoding~\cite{shi2024denoising} and high-norm outlier tokens associated with implicit memory storage~\cite{darcet2023vision}.
Register tokens provide dedicated storage slots for such outlier behavior~\cite{darcet2023vision, lappe2025register}, which is also related to attention sink phenomena in language models~\cite{barbero2025llm}.
Specialized processing of class and patch tokens further improves local representations~\cite{marouani2026revisiting}.
In NVS transformers, however, each patch token is also bound to a camera ray, so feature artifacts are coupled with semantic-spatial mixing rather than token dynamics alone.
This motivates us to separate semantic and spatial representations instead of relying only on additional register tokens.

\textbf{Representation Alignment.}
The DINO series~\cite{caron2021dino, oquab2023dinov2, simeoni2025dinov3} has shown that self-supervised ViTs learn useful visual representations.
REPA~\cite{yu2024representation} and its variants~\cite{tian2025urepa, leng2025repae} align hidden states with such features to improve generative training, while iREPA~\cite{singh2025iREPA} emphasizes fine-grained spatial structure.
Representation autoencoders~\cite{zheng2025diffusion} instead use pretrained visual features as diffusion latents.
Applying this supervision to unified NVS tokens can mix semantic and geometric objectives.
We therefore use iREPA with DINOv3 as branch-specific supervision for the decoupled semantic stream.

\textbf{Geometric Priors for 3D Understanding.}
Monocular and multi-view depth estimation has matured into a family of large-scale foundation models~\cite{bhat2023zoedepth, yang2024depthanything, yang2024depthanythingv2, bochkovskii2024depthpro, wang2024moge, wang2024dust3r}.
DA3~\cite{lin2025depthanything3} extends this line to multi-view consistent geometry prediction through depth and rays.
When geometric supervision is enabled, we use its outputs to supervise the decoupled spatial stream during training, without adding an inference-time geometry module.
LagerNVS~\cite{lagernvs} instead initializes its inference-time encoder from a geometry-pretrained reconstruction network.

\begin{figure}[!t]
    \centering
    \includegraphics[width=.98\linewidth]
    {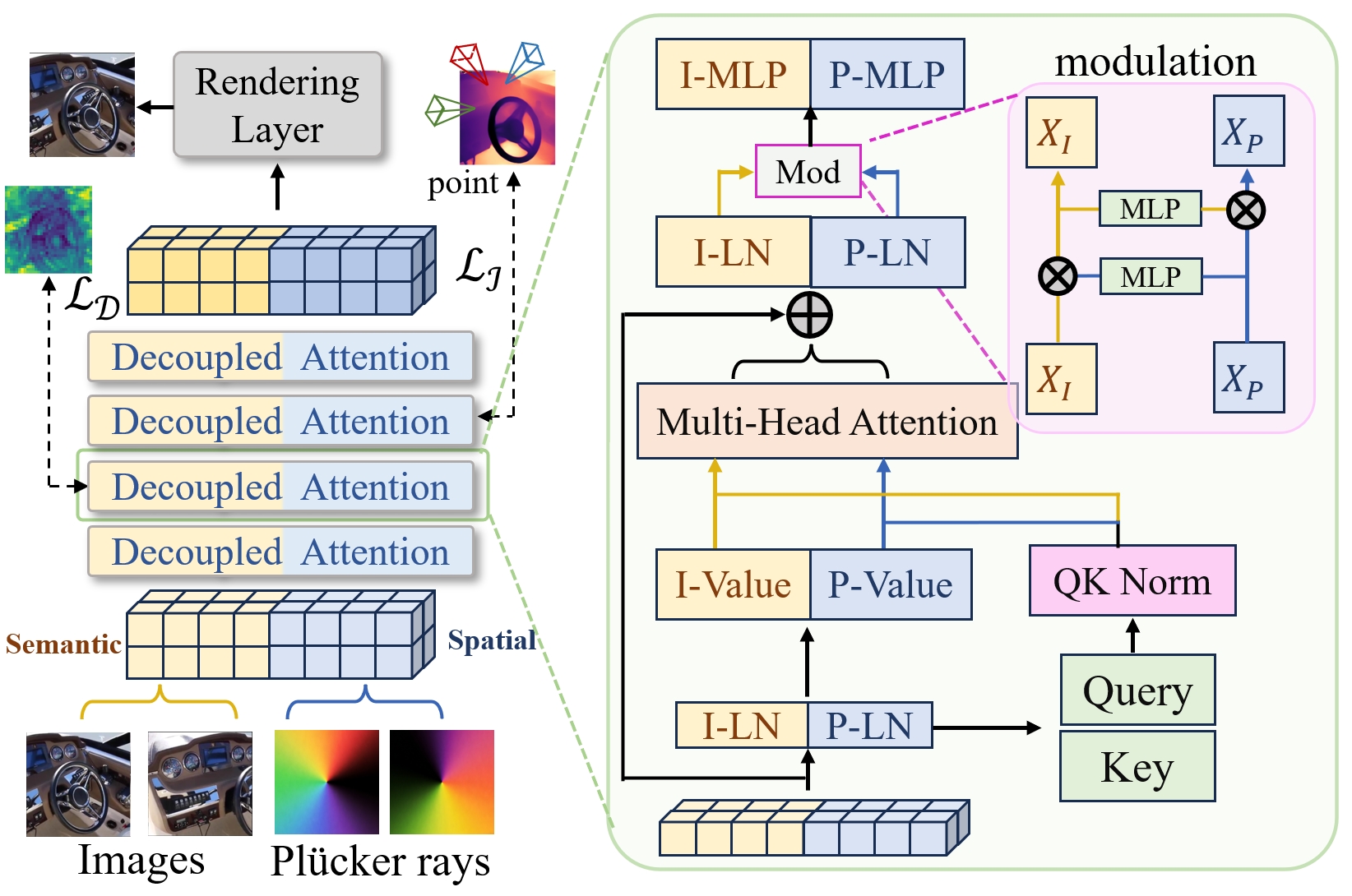}
    \vspace{-2mm}
    \caption{\textbf{Semantic--Spatial Decoupled Architecture for Feedforward NVS.}
    Semantic tokens ($I$) from RGB and spatial tokens ($P$) from Pl\"ucker rays pass through decoupled-attention blocks.
    Attention shares query--key interactions while using independent value projections to preserve heterogeneous representations.
    Bidirectional modulation further enables cross-stream conditioning.}

    \vspace{-7mm}
    \label{fig:model}
\end{figure}

\section{Method}
\label{sec:method}

\subsection{Preliminary}
\label{sec:preliminary}

Given $N$ posed input images $\{(\mathbf{I}_i,\mathbf{E}_i,\mathbf{K}_i)\}_{i=1}^{N}$ and
$M$ target poses $\{(\mathbf{E}_j,\mathbf{K}_j)\}_{j=1}^{M}$, feedforward NVS renders
$\{\hat{\mathbf{I}}_j\}_{j=1}^{M}$ in a single forward pass, where $\mathbf{I}_i\in\mathbb{R}^{H\times W\times 3}$
is the input RGB image, $\mathbf{E}_i,\mathbf{K}_i\in\mathbb{R}^{4\times 4}$ are the extrinsic
and intrinsic matrices, and $\hat{\mathbf{I}}_j\in\mathbb{R}^{H\times W\times 3}$ is the rendered
output. In this work, we adopt LVSM~\cite{jin2024lvsm} as the baseline. The camera viewpoints are encoded as Pl\"ucker ray
maps~\cite{plucker1865neue} $\boldsymbol{\pi}_i\in\mathbb{R}^{6\times H\times W}$.
Each input view is divided into $N_p=HW/p^2$ non-overlapping patches of size $p{\times}p$;
the RGB and Pl\"ucker patch sequences, denoted $\mathbf{I}_i^\mathbf{p}$ and
$\boldsymbol{\pi}_i^\mathbf{p}$, are \emph{concatenated along the channel dimension} and
projected into the $D$-dimensional token space, while target tokens are
Pl\"ucker patches alone:
\begin{equation}\setlength{\abovedisplayskip}{2pt} \setlength{\belowdisplayskip}{2pt} \setlength{\abovedisplayshortskip}{2pt} \setlength{\belowdisplayshortskip}{2pt}
  \mathbf{t}_i = \text{MLP}_{\text{in}}(\operatorname{concat}(\mathbf{I}_i^\mathbf{p},
  \boldsymbol{\pi}_i^\mathbf{p})) \in \mathbb{R}^{N_p \times D},
  \qquad
  \mathbf{q}_j = \text{MLP}_{\text{tgt}}(\boldsymbol{\pi}_j^\mathbf{p})
  \in \mathbb{R}^{N_p \times D}.
\end{equation}
All tokens are then concatenated and processed by $L$ layers of full self-attention,
after which target features are fed into the rendering head.

\subsection{Decoupled Semantic and Spatial Tokens}
\label{sec:decouple}

\noindent\textbf{Decoupled Tokenization.}
As shown in Fig.~\ref{fig:model}, we replace the modality-mixing input projection in LVSM with two separate tokenizers for semantic appearance and spatial geometry. For each input view, RGB patches $\mathbf{I}_i^\mathbf{p}$ and Pl\"ucker-ray patches $\boldsymbol{\pi}_i^\mathbf{p}$ are tokenized into a semantic branch ($I$-branch) and a spatial branch ($P$-branch):
\begin{equation}\setlength{\abovedisplayskip}{2pt} \setlength{\belowdisplayskip}{2pt} \setlength{\abovedisplayshortskip}{2pt} \setlength{\belowdisplayshortskip}{2pt}
  \mathbf{x}_i^I = \text{Tok}_I(\mathbf{I}_i^\mathbf{p})\in\mathbb{R}^{N_p\times \frac{D}{2}},
  \qquad
  \mathbf{x}_i^P = \text{Tok}_P(\boldsymbol{\pi}_i^\mathbf{p})\in\mathbb{R}^{N_p\times \frac{D}{2}} .
\end{equation}
For each target view, no RGB observation is available, so we initialize the target semantic branch with zeros and tokenize the target Pl\"ucker-ray patches into the spatial branch:
\begin{equation}\setlength{\abovedisplayskip}{2pt} \setlength{\belowdisplayskip}{2pt} \setlength{\abovedisplayshortskip}{2pt} \setlength{\belowdisplayshortskip}{2pt}
  \mathbf{y}_j^I = \mathbf{0}\in\mathbb{R}^{N_p\times \frac{D}{2}},
  \qquad
  \mathbf{y}_j^P = \text{Tok}_{P}^{\text{tgt}}(\boldsymbol{\pi}_j^\mathbf{p})
  \in\mathbb{R}^{N_p\times \frac{D}{2}} .
\end{equation}
The full input and target tokens are then formed by concatenating their $I$-branch and $P$-branch features. The total token dimension therefore remains the same as the original LVSM token, while the branch partition is preserved throughout the Transformer blocks.

\noindent\textbf{Independent-V Attention.}
Each decoupled Transformer block keeps query-key (Q/K) routing on the full token, but uses branch-specific value and output projections. With branch-wise pre-normalized inputs $\bar{\mathbf{x}}^I$ and $\bar{\mathbf{x}}^P$, let $\bar{\mathbf{x}}=\operatorname{concat}(\bar{\mathbf{x}}^I,\bar{\mathbf{x}}^P)$. We compute
\begin{equation}\setlength{\abovedisplayskip}{2pt} \setlength{\belowdisplayskip}{2pt} \setlength{\abovedisplayshortskip}{2pt} \setlength{\belowdisplayshortskip}{2pt}
  \mathbf{q}=\bar{\mathbf{x}}\mathbf{W}_q,\qquad
  \mathbf{k}=\bar{\mathbf{x}}\mathbf{W}_k,\qquad
  \mathbf{v}^I=\bar{\mathbf{x}}^I\mathbf{W}_v^I,\quad
  \mathbf{v}^P=\bar{\mathbf{x}}^P\mathbf{W}_v^P .
\end{equation}
Here $\mathbf{W}_q$ and $\mathbf{W}_k$ project the full token to Q/K, while $\mathbf{W}_v^I$ and $\mathbf{W}_v^P$ project the two value streams separately. The shared attention map is
\begin{equation}\setlength{\abovedisplayskip}{2pt} \setlength{\belowdisplayskip}{2pt} \setlength{\abovedisplayshortskip}{2pt} \setlength{\belowdisplayshortskip}{2pt}
  \mathbf{A}=\operatorname{softmax}\!\left(\frac{\mathbf{q}\mathbf{k}^{\top}}{\sqrt{d_h}}\right),
\end{equation}
where $d_h$ is the per-head Q/K dimension. The same attention map is applied to the two branch-specific value streams and followed by branch-specific output projections:
\begin{equation}\setlength{\abovedisplayskip}{2pt} \setlength{\belowdisplayskip}{2pt} \setlength{\abovedisplayshortskip}{2pt} \setlength{\belowdisplayshortskip}{2pt}
  \mathbf{o}=\operatorname{concat}\!\left((\mathbf{A}\mathbf{v}^I)\mathbf{W}_o^I,
  (\mathbf{A}\mathbf{v}^P)\mathbf{W}_o^P\right).
\end{equation}

where $\mathbf{W}_o^I$ and $\mathbf{W}_o^P$ are the output projections for the two branches. After attention, the $I$-branch and $P$-branch are processed by branch-specific layer normalization (LN) and half-dimensional feed-forward networks (FFNs), then concatenated back to a $D$-dimensional token. Thus, the block preserves full-token interaction through shared Q/K while keeping value updates and FFN transformations branch-specific. The qualitative feature progression in Fig.~\ref{fig:compare} visualizes the resulting change in intermediate representations, with detailed analysis provided in Sec.~\ref{sec:analysis_artifacts}.

\subsection{Optional Categorized Supervision}
\label{sec:supervision}

The decoupled representation enables optional categorized supervision for the $I$-branch and $P$-branch. 

\noindent\textbf{Semantic Alignment via iREPA with DINOv3.}
For the $I$-branch, we align intermediate semantic features with frozen DINOv3 features using an iREPA-style representation alignment~\cite{singh2025iREPA}. Let $\mathbf{x}_{k_I}^I$ denote the $I$-branch feature at layer $k_I$, $f_{\text{DINO}}(\mathbf{I})$ denote the DINOv3 teacher feature, $\operatorname{SN}(\cdot)$ denote spatial normalization, and $h_\phi$ denote the semantic projection head. The semantic supervision is
\begin{equation}\setlength{\abovedisplayskip}{2pt} \setlength{\belowdisplayskip}{2pt} \setlength{\abovedisplayshortskip}{2pt} \setlength{\belowdisplayshortskip}{2pt}
  \mathcal{L}_{\text{iREPA}}
  =
  \operatorname{SmoothL1}\!\left(
  h_\phi(\operatorname{SN}(\mathbf{x}_{k_I}^I)),
  f_{\text{DINO}}(\mathbf{I})
  \right).
\end{equation}
For a token feature $\mathbf{z}$, spatial normalization is defined as
\begin{equation}\setlength{\abovedisplayskip}{2pt} \setlength{\belowdisplayskip}{2pt} \setlength{\abovedisplayshortskip}{2pt} \setlength{\belowdisplayshortskip}{2pt}
  \operatorname{SN}(\mathbf{z})
  =
  \frac{\mathbf{z}-\gamma\mu(\mathbf{z})}
  {\sigma(\mathbf{z})+\epsilon},
\end{equation}
where $\mu(\mathbf{z})$ and $\sigma(\mathbf{z})$ are computed over the token dimension, $\gamma$ is a normalization coefficient, and $\epsilon$ is a small constant for numerical stability.
Inspired by iREPA, this normalization emphasizes local feature variation before alignment. Both $h_\phi$ and the DINOv3 teacher are discarded at inference.

\vspace{-1mm}
\noindent\textbf{Geometric Consistency of the Spatial Branch.}
For the $P$-branch, we use DA3-derived geometry~\cite{lin2025depthanything3} to construct visible cross-view correspondences. Specifically, source-view 3D points are projected into target views with the aligned camera parameters; when multiple source points map to the same target location, z-buffering keeps the point with the smallest target-view depth and filters occluded matches. Let $\Omega_j$ be the resulting correspondence set for target view $j$, where $(i,u,v)\in\Omega_j$ matches source location $u$ in view $i$ to target location $v$ in view $j$. We supervise the spatial branch with feature cosine consistency:
\begin{equation}\setlength{\abovedisplayskip}{2pt} \setlength{\belowdisplayskip}{2pt} \setlength{\abovedisplayshortskip}{2pt} \setlength{\belowdisplayshortskip}{2pt}
  \mathcal{L}_{\text{geo}}
  =
  \frac{1}{|\mathcal{J}|}
  \sum_{j\in\mathcal{J}}
  \frac{1}{|\Omega_j|}
  \sum_{(i,u,v)\in\Omega_j}
  \left(
  1 -
  \cos\!\left(\mathbf{x}_{k_P,i,u}^{P}, \mathbf{x}_{k_P,j,v}^{P}\right)
  \right),
\end{equation}
where $\mathcal{J}$ denotes target views with valid correspondences, $\mathbf{x}_{k_P,i,u}^{P}$ and $\mathbf{x}_{k_P,j,v}^{P}$ denote $P$-branch features at layer $k_P$ for the matched locations, and $\cos(\cdot,\cdot)$ denotes cosine similarity. The cached DA3 geometry is used to build correspondence supervision during training and is not used at inference.

\subsection{Optional Bidirectional Feature Modulation}
\label{sec:film}

Bidirectional modulation is an optional refinement on top of the decoupled Transformer block. It uses cross-branch conditioning~\cite{perez2018film} after layer normalization and before the branch-specific FFNs. 

\noindent\textbf{Spatial $\rightarrow$ Semantic.}
Let $\mathbf{m}^I$ and $\mathbf{m}^P$ denote the layer-normalized attention outputs
of the I and P branches respectively, \ie, the inputs to their corresponding FFNs.
The spatial representation $\mathbf{m}^P$ generates scale and shift parameters that
modulate the semantic branch:
\begin{equation}\setlength{\abovedisplayskip}{2pt} \setlength{\belowdisplayskip}{2pt} \setlength{\abovedisplayshortskip}{2pt} \setlength{\belowdisplayshortskip}{2pt}
  (\boldsymbol{\gamma}^{P\to I},\boldsymbol{\beta}^{P\to I})
  = \mathbf{m}^P\mathbf{W}_{\mathrm{F}}^{P}+\mathbf{b}_{\mathrm{F}}^{P},
  \qquad
  \tilde{\mathbf{m}}^I
  = \boldsymbol{\gamma}^{P\to I}\odot\mathbf{m}^I
  + \boldsymbol{\beta}^{P\to I}.
\end{equation}
This step lets spatial cues such as view direction and ray geometry condition the semantic update without merging spatial channels into the semantic representation.

\noindent\textbf{Semantic $\rightarrow$ Spatial.}
The modulated semantic representation $\tilde{\mathbf{m}}^I$ then generates the reverse affine parameters for the spatial branch:
\begin{equation}\setlength{\abovedisplayskip}{2pt} \setlength{\belowdisplayskip}{2pt} \setlength{\abovedisplayshortskip}{2pt} \setlength{\belowdisplayshortskip}{2pt}
  (\boldsymbol{\gamma}^{I\to P},\boldsymbol{\beta}^{I\to P})
  = \tilde{\mathbf{m}}^I\mathbf{W}_{\mathrm{F}}^{I}+\mathbf{b}_{\mathrm{F}}^{I},
  \qquad
  \tilde{\mathbf{m}}^P
  = \boldsymbol{\gamma}^{I\to P}\odot\mathbf{m}^P
  + \boldsymbol{\beta}^{I\to P}.
\end{equation}
The branch-specific FFNs are then applied to $\tilde{\mathbf{m}}^I$ and $\tilde{\mathbf{m}}^P$ separately. Thus, modulation provides controlled bidirectional conditioning while preserving the decoupled representation.

Here $\mathbf{W}_{\mathrm{F}}^{P}$ and $\mathbf{W}_{\mathrm{F}}^{I}$ are learnable generator weights, $\mathbf{b}_{\mathrm{F}}^{P}$ and $\mathbf{b}_{\mathrm{F}}^{I}$ are their biases, and $\boldsymbol{\gamma},\boldsymbol{\beta}$ are the predicted scale and shift parameters. We initialize the generators as an identity modulation, with zero weights, scale bias initialized to 1, and shift bias initialized to 0. Therefore, enabling modulation starts from the original decoupled block and learns cross-branch corrections when they are useful.

\subsection{Training Objective}
\label{sec:loss}

The base decoupled model is trained with the RGB reconstruction loss. When optional categorized supervision is enabled, the objective becomes
\begin{equation}\setlength{\abovedisplayskip}{2pt} \setlength{\belowdisplayskip}{2pt} \setlength{\abovedisplayshortskip}{2pt} \setlength{\belowdisplayshortskip}{2pt}
  \mathcal{L}
  =
  \mathcal{L}_{\text{RGB}}
  + \lambda_I \mathcal{L}_{\text{iREPA}}
  + \lambda_P \mathcal{L}_{\text{geo}} .
\end{equation}
Here $\mathcal{L}_{\text{iREPA}}$ and $\mathcal{L}_{\text{geo}}$ constrain the $I$-branch and $P$-branch only during training. Implementation details and loss weights are provided in Appendix~\ref{sec:appendix_impl}.

\section{Experiments}
\label{sec:experiments}

\subsection{Datasets}

\noindent\textbf{Scene-Level.}
For scene-level, we use RealEstate10K~\cite{zhou2018stereo}, 
which aggregates over 80K video clips of diverse indoor and outdoor environments 
collected from YouTube. The dataset provides large-scale real-world scenes with 
varied viewpoints and camera motions. Train and test splits follow LVSM~\cite{jin2024lvsm}.

\noindent\textbf{Object-Level.}
Object-level experiments use Objaverse~\cite{deitke2023objaverse} for
training, with 730K objects each rendered at 32 viewpoints using the
GS-LRM~\cite{zhang2024gslrm} pipeline.
Evaluation is performed on the Objaverse test split.
At test time, each object is presented with 4 structured input views,
and performance is measured on 10 held-out target views sampled at random,
following the protocol of LVSM~\cite{jin2024lvsm}.

\subsection{Implementation Details}
\label{sec:impl}

\noindent\textbf{Model Configurations.}
Our main experiments use a 12-layer LVSM Decoder-Only~\cite{jin2024lvsm}
and an LVSM Encoder-Decoder with a 12-layer encoder and 12-layer decoder;
both use a hidden dimension of 768. We also test a 24-layer Decoder-Only backbone.

\noindent\textbf{Controlled Reimplementation.}
Our LVSM baselines and variants share the codebase, data splits, view sampling,
resolution, and training budget. We train for 50K iterations, using 2 input
and 6 target views at $256{\times}256$ on RealEstate10K, and 4 input and
8 target views on Objaverse. External reconstruction methods are evaluated
on the same benchmarks but retain different architectures, camera encodings,
and pretrained priors. Optimizer, loss, and supervision details are collected
in Appendix~\ref{sec:appendix_impl}.
We report PSNR, SSIM~\cite{wang2004image}, and LPIPS~\cite{zhang2018lpips} for all quantitative comparisons.

\begin{table}[t]
\centering
\caption{\textbf{Reconstruction and camera-encoding comparisons.} Full: decoupling with optional supervision and modulation. Params in millions; Enc-Dec: encoder--decoder; 24L: 24-layer decoder-only; P/R: Pl\"ucker rays/raw raymaps; PRoPE adds projective camera conditioning.}
\label{tab:main}
\scriptsize
\setlength{\tabcolsep}{1.8pt}
\makebox[\textwidth][c]{%
\begin{minipage}[t]{0.50\textwidth}
\vspace{0pt}
\centering
\textbf{(a) RealEstate10K reconstruction}\\
\renewcommand{\arraystretch}{1.13}
\begin{tabularx}{\linewidth}{@{}>{\hsize=2\hsize\raggedright\arraybackslash}X*{4}{>{\hsize=.75\hsize\raggedleft\arraybackslash}X}@{}}
\toprule
\textbf{Method} & \textbf{Params} & \textbf{PSNR}$\uparrow$ & \textbf{SSIM}$\uparrow$ & \textbf{LPIPS}$\downarrow$ \\
\midrule
LagerNVS-DA3-S/4~\cite{lagernvs} & 55.3 & 24.91 & 0.801 & 0.146 \\
MVSplat~\cite{mvsplat} & 11.9 & 25.48 & 0.851 & 0.143 \\
PRoPE~\cite{prope} & 85.8 & 26.79 & 0.851 & 0.136 \\
DepthSplat~\cite{depthsplat} & 120.7 & 26.43 & 0.865 & 0.127 \\
pixelSplat~\cite{pixelsplat} & 119.0 & 25.83 & 0.859 & 0.138 \\
\rowcolor{slateL}
LVSM~\cite{jin2024lvsm} & 85.8 & 26.10 & 0.839 & 0.144 \\
\rowcolor{slateM}
LVSM Decouple & 50.1 & 26.70 & 0.851 & 0.138 \\
\rowcolor{slateH}
\textbf{Ours (full)} & 57.2 & \emph{27.21} & \textbf{0.869} & \emph{0.125} \\
\midrule
\rowcolor{slateL}
LVSM (24L) & --- & 26.18 & 0.8420 & 0.1406 \\
\rowcolor{slateH}
\textbf{Ours (24L)} & --- & \textbf{27.37} & \emph{0.8671} & \textbf{0.1249} \\
\midrule
\rowcolor{slateL}
LVSM (Enc-Dec) & --- & 24.06 & 0.775 & 0.206 \\
\rowcolor{slateM}
Decouple (Enc-Dec) & --- & 24.84 & 0.789 & 0.170 \\
\rowcolor{slateH}
\textbf{Ours (Enc-Dec, full)} & --- & \textbf{25.31} & \textbf{0.806} & \textbf{0.154} \\
\bottomrule
\end{tabularx}
\end{minipage}%
\hspace{0.02\textwidth}%
\begin{minipage}[t]{0.48\textwidth}
\vspace{0pt}
\centering
\textbf{(b) Objaverse}\\
\renewcommand{\arraystretch}{1.13}
\begin{tabularx}{\linewidth}{@{}>{\hsize=2\hsize\raggedright\arraybackslash}X*{4}{>{\hsize=.75\hsize\raggedleft\arraybackslash}X}@{}}
\toprule
\textbf{Method} & \textbf{Params} & \textbf{PSNR}$\uparrow$ & \textbf{SSIM}$\uparrow$ & \textbf{LPIPS}$\downarrow$ \\
\midrule
LGM~\cite{lgm} & 56.980 & 22.61 & 0.834 & 0.193 \\
\rowcolor{slateL}
LVSM~\cite{jin2024lvsm} & 85.841 & 23.75 & 0.864 & 0.150 \\
\rowcolor{slateM}
LVSM Decouple & 50.083 & \emph{25.96} & \emph{0.880} & \emph{0.112} \\
\rowcolor{slateH}
\textbf{Ours (full)} & 57.188 & \textbf{26.46} & \textbf{0.899} & \textbf{0.101} \\
\bottomrule
\end{tabularx}

\vspace{1.5mm}
\textbf{(c) Camera encoding (RealEstate10K)}\\
\begin{tabularx}{\linewidth}{@{}>{\hsize=2\hsize\raggedright\arraybackslash}X*{4}{>{\hsize=.75\hsize\raggedleft\arraybackslash}X}@{}}
\toprule
\textbf{Method} & \textbf{Params} & \textbf{PSNR}$\uparrow$ & \textbf{SSIM}$\uparrow$ & \textbf{LPIPS}$\downarrow$ \\
\midrule
PRoPE (P)~\cite{prope} & 85.8 & 26.79 & 0.851 & \emph{0.136} \\
\rowcolor{slateL}
LVSM (P) & 85.8 & 26.10 & 0.839 & 0.144 \\
\rowcolor{slateL}
LVSM (R) & 85.8 & 25.78 & 0.829 & 0.149 \\
\rowcolor{slateM}
Decouple (R) & 50.1 & \emph{26.82} & \emph{0.853} & 0.138 \\
\rowcolor{slateM}
Decouple (P) & 50.1 & 26.70 & 0.851 & 0.138 \\
\rowcolor{slateH}
\textbf{Ours (P)} & 57.2 & \textbf{27.21} & \textbf{0.869} & \textbf{0.125} \\
\bottomrule
\end{tabularx}
\end{minipage}%
}
\vspace{-5mm}
\end{table}

\begin{figure}[!t]
    \centering
    \includegraphics[width=.98\linewidth]
    {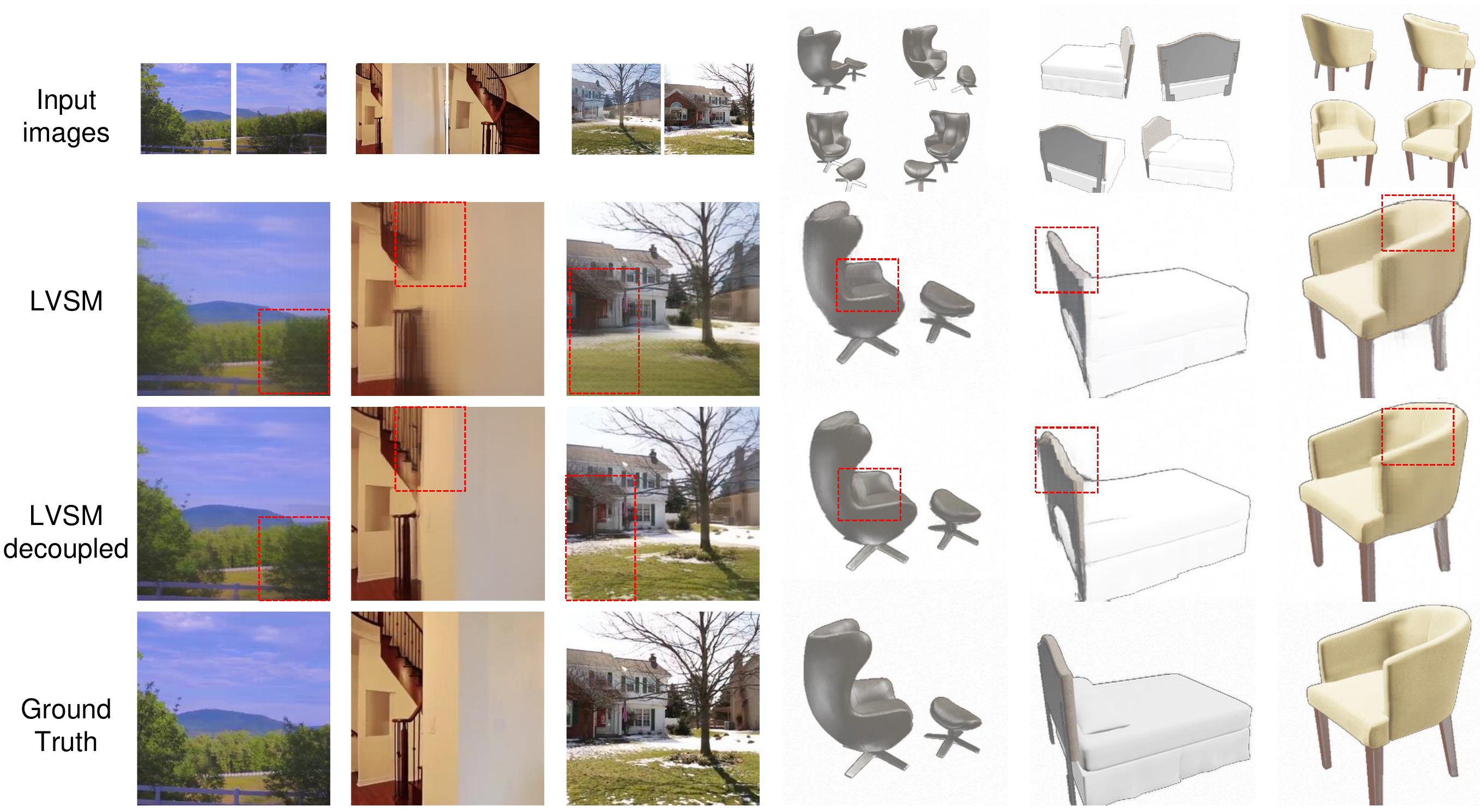}
    \vspace{-1mm}
    \caption{\textbf{Novel View Synthesis Visual Comparison.}
    Our decoupled model produces more coherent structures and sharper details.}
    \vspace{-5mm}
    \label{fig:infer_compare}
\end{figure}

\begin{table}[t]
\centering
\caption{\textbf{Component Ablation.}
Reg: register tokens. ${}^{\dagger}$Training diverged: result unavailable.}
\label{tab:ablation_main}

\scriptsize
\renewcommand{\arraystretch}{1.14}
\setlength{\tabcolsep}{2.2pt}

\makebox[\textwidth][c]{%
\begin{minipage}[t]{0.485\textwidth}
\vspace{0pt}
\centering
\begin{tabularx}{\linewidth}{@{}>{\raggedright\arraybackslash}Xccc@{}}
\toprule
\textbf{Configuration} & \textbf{PSNR}$\uparrow$ & \textbf{SSIM}$\uparrow$ & \textbf{LPIPS}$\downarrow$ \\
\midrule

\rowcolor{slateL!45}
\multicolumn{4}{@{}l@{}}{\textbf{Decoupling and Categorized Supervision}} \\
\textbf{Baseline}
& \textbf{26.10} & \textbf{0.839} & \textbf{0.144} \\
Baseline + Sup
& 26.02 & 0.831 & 0.148 \\
\rowcolor{slateL}
\textbf{Decouple}
& \textbf{26.70} & \textbf{0.851} & \textbf{0.138} \\
\rowcolor{slateM}
\textbf{Decouple + Sup}
& \textbf{26.91} & \textbf{0.857} & \textbf{0.134} \\

\midrule

\rowcolor{slateL!45}
\multicolumn{4}{@{}l@{}}{\textbf{Register Control}} \\
Baseline + Reg
& 26.04 & 0.835 & 0.146 \\
\textbf{Baseline}
& \textbf{26.10} & \textbf{0.839} & \textbf{0.144} \\

\midrule

\rowcolor{slateL!45}
\multicolumn{4}{@{}l@{}}{\textbf{Shared vs.\ Independent Q/K}} \\
Independent-Q/K
& 25.90 & 0.832 & 0.148 \\
\rowcolor{slateL}
\textbf{Shared-Q/K}
& \textbf{26.70} & \textbf{0.851} & \textbf{0.138} \\

\midrule

\rowcolor{slateL!45}
\multicolumn{4}{@{}l@{}}{\textbf{Modulation Placement}} \\
Pre-Attn-LN$^{\dagger}$
& --- & --- & --- \\
Post-FFN
& 26.95 & 0.858 & 0.132 \\
\rowcolor{slateH!65}
\textbf{Post-Attn-LN/Pre-FFN}
& \textbf{27.06} & \textbf{0.860} & \textbf{0.131} \\

\bottomrule
\end{tabularx}
\end{minipage}%
\hspace{0.025\textwidth}%
\begin{minipage}[t]{0.485\textwidth}
\vspace{0pt}
\centering
\renewcommand{\arraystretch}{1.220}
\begin{tabularx}{\linewidth}{@{}>{\raggedright\arraybackslash}Xccc@{}}
\toprule
\textbf{Configuration} & \textbf{PSNR}$\uparrow$ & \textbf{SSIM}$\uparrow$ & \textbf{LPIPS}$\downarrow$ \\
\midrule

\rowcolor{slateL!45}
\multicolumn{4}{@{}l@{}}{\textbf{Effect of Bidirectional Modulation}} \\
\textbf{Baseline}
& \textbf{26.10} & \textbf{0.839} & \textbf{0.144} \\
Baseline + Mod
& 25.93 & 0.829 & 0.150 \\
\rowcolor{slateL}
\textbf{Decouple}
& \textbf{26.70} & \textbf{0.851} & \textbf{0.138} \\
\rowcolor{slateH!65}
\textbf{Decouple + Mod}
& \textbf{27.06} & \textbf{0.860} & \textbf{0.131} \\

\midrule

\rowcolor{slateL!45}
\multicolumn{4}{@{}l@{}}{\textbf{Cross-Branch Modulation Direction}} \\
I$\to$P only
& 26.80 & 0.853 & 0.136 \\
P$\to$I only
& 26.91 & 0.856 & 0.133 \\
\rowcolor{slateH!65}
\textbf{Decouple + Mod}
& \textbf{27.06} & \textbf{0.860} & \textbf{0.131} \\

\midrule

\rowcolor{slateL!45}
\multicolumn{4}{@{}l@{}}{\textbf{LayerNorm Decoupling}} \\
Joint-LN
& 26.01 & 0.830 & 0.149 \\
\rowcolor{slateL}
\textbf{Decouple}
& \textbf{26.70} & \textbf{0.851} & \textbf{0.138} \\

\midrule

\rowcolor{slateL!45}
\multicolumn{4}{@{}l@{}}{\textbf{Full Configuration}} \\
\rowcolor{slateH}
\textbf{Decouple + Sup + Mod}
& \textbf{27.21} & \textbf{0.869} & \textbf{0.125} \\

\bottomrule
\end{tabularx}
\renewcommand{\arraystretch}{1.14}
\end{minipage}%
}

\vspace{-3mm}
\end{table}

\subsection{Comparison with Baselines}
\label{sec:main_results}

Table~\ref{tab:main} compares our variants with recent reconstruction methods under common training and evaluation protocols. We match data splits, view sampling, resolution, batch size, training budget, and evaluation views.
Decoupling improves both LVSM Decoder-Only
and Encoder--Decoder backbones over their corresponding baselines;
optional supervision and modulation bring further gains in the full variants.
Improvements also hold with the 24-layer Decoder-Only backbone and raw
raymaps. Fig.~\ref{fig:infer_compare} shows the visual improvement; Detailed comparison settings are provided in Appendix~\ref{sec:appendix_impl}.
Additional \href{https://anonymous.4open.science/w/supplement-for-rebuttal-47B1/index.html#qualitative}{qualitative comparisons}
on both datasets and RealEstate10K
\href{https://anonymous.4open.science/w/supplement-for-rebuttal-47B1/index.html#video-results}{rendering videos} are provided online.

\subsection{Ablation Study}
\label{sec:analysis_artifacts}

Table.~\ref{tab:ablation_main} summarizes the component and control ablations. Unless otherwise specified, feature visualizations are taken from middle Transformer layers.

\begin{figure}[!t]
    \centering
    \includegraphics[width=.95\linewidth]
    {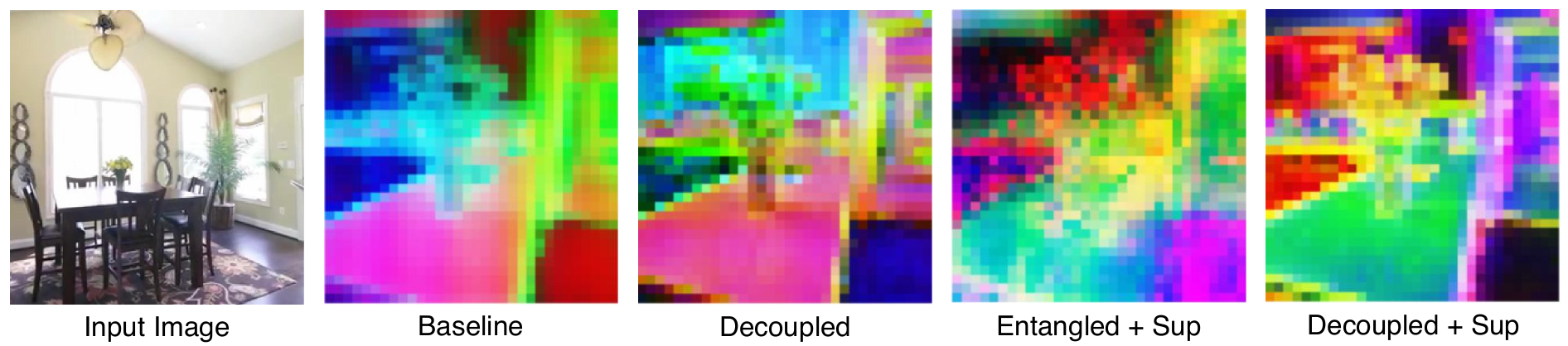}
    \vspace{-2mm}
    \caption{\textbf{Supervision Difference.} Categorized supervision benefits the decoupled model but harms the entangled baseline.} 
    \vspace{-7mm}
    \label{fig:sup+decouple}
\end{figure}

\noindent\textbf{Decoupling and Categorized Supervision.}
In the entangled baseline, RGB and Pl\"ucker information share the same feature channels, so semantic alignment and geometric consistency are imposed on a mixed representation and do not improve rendering quality. After decoupling, the two objectives become well-defined: iREPA constrains the $I$-branch, while geometric correspondence constrains the $P$-branch. Fig.~\ref{fig:sup+decouple} shows clearer semantic and geometric structures, confirming that supervision is secondary to, and enabled by, the decoupled representation.

\begin{figure}[!t]
    \centering
    \begin{subfigure}[t]{0.54\linewidth}
        \centering
        \includegraphics[width=\linewidth]{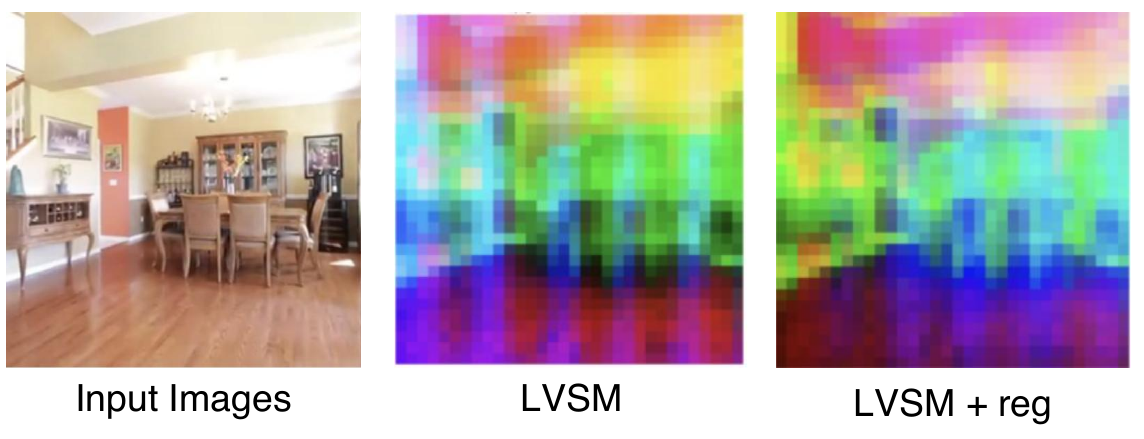}
        \vspace{-4mm}
        \caption{\textbf{Register Tokens Ablation.}
        Register tokens alone do not remove the grid-like artifacts.}
        \label{fig:reg}
    \end{subfigure}
    \hfill
    \begin{subfigure}[t]{0.40\linewidth}
        \centering
        \includegraphics[width=\linewidth]{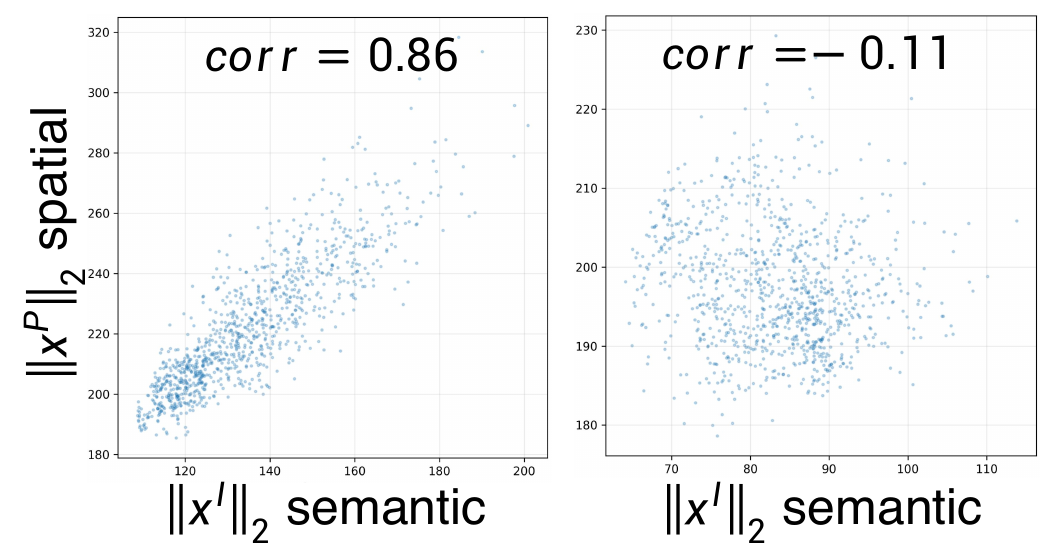}
        \vspace{-4mm}
        \caption{\textbf{LayerNorm Ablation.}
        Entangled representations exhibit strong norm correlation.}
        \label{fig:scatter}
    \end{subfigure}
    \vspace{-4mm}
\end{figure}

\noindent\textbf{Register Control.}
We add register tokens to the entangled baseline under the same setting to test
whether the grid-like artifacts can be handled by providing extra storage slots.
This control does not improve rendering quality, and the grid-like artifacts remain visible in the feature maps (Fig.~\ref{fig:reg}). This suggests that the artifacts are not fully explained by the lack of extra storage slots.
GAS measurements and controlled feature interventions examine
grid structure and its effect on rendering (Appendix~\ref{sec:appendix_artifacts}).

\begin{figure}[!t]
    \centering
    \includegraphics[width=.95\linewidth]
    {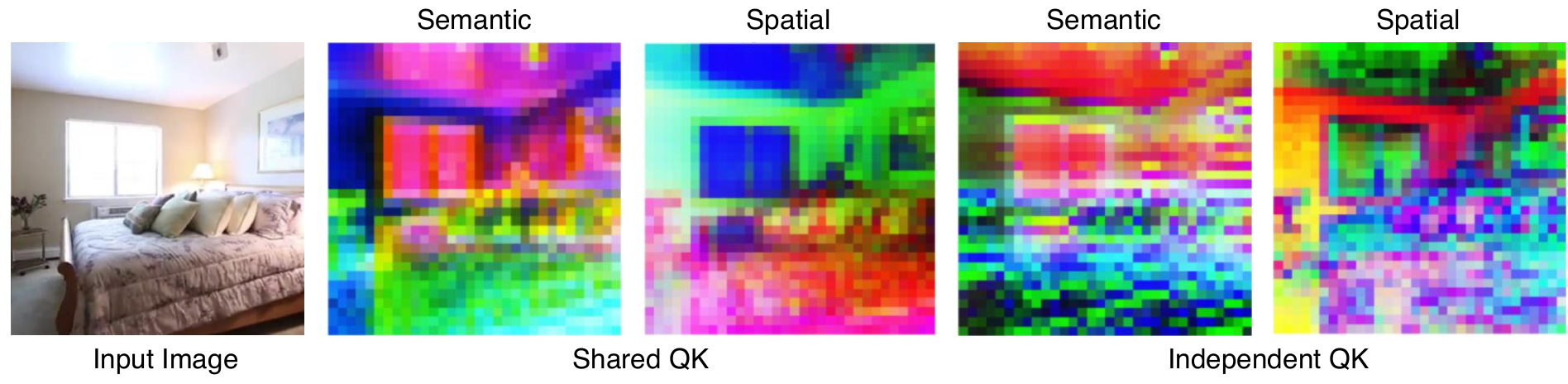}
    \vspace{-2mm}\caption{\textbf{Shared vs.\ Independent QK.} Shared Q/K with independent V yields better performance.}
    \vspace{-6mm}
    \label{fig:qk}
\end{figure}

\begin{figure}[!t]
    \centering
    \includegraphics[width=.95\linewidth]{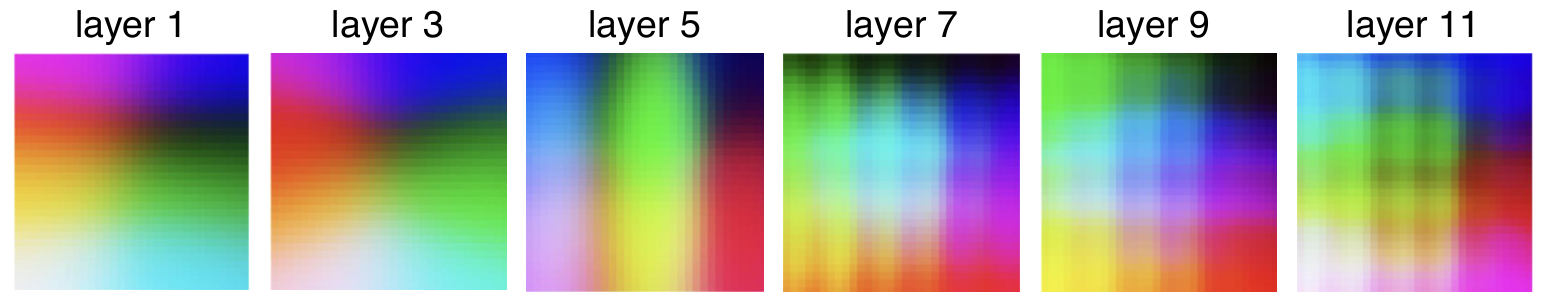}
    \vspace{-2mm}
    \caption{\textbf{Pl\"ucker-Only Inference Control.}
    Artifacts originate from the spatial ray representation.}
    \vspace{-8mm}
    \label{fig:plucker_control}
\end{figure}

\noindent\textbf{Pl\"ucker-Only Inference Control.}
To further localize the source of the grid-like artifacts, we run the decoupled model at inference time with all input RGB values set to zero while keeping the Pl\"ucker rays unchanged. As shown in Fig.~\ref{fig:plucker_control}, the intermediate features still exhibit clear network-like grid patterns. Since image appearance has been removed in this control, the remaining artifacts provide evidence that the grid-like patterns are induced by the spatial Pl\"ucker-ray representation.

\noindent\textbf{Alternative Camera Encodings.}
We replace Pl\"ucker rays $(o\times d,d)$ with raw raymaps $(o,d)$, where $o$ and $d$ denote the ray origin and direction, while keeping the LVSM architecture and training protocol otherwise unchanged. Panel (c) of Table~\ref{tab:main} shows that decoupling improves all three rendering metrics with raw raymaps, indicating that its benefit is not specific to the Pl\"ucker parameterization. PRoPE~\cite{prope} incorporates projective camera geometry into attention, whereas our design addresses the entanglement caused by jointly propagating semantic and spatial features. This distinction positions decoupling as a representation-level design complementary to camera conditioning.

\noindent\textbf{Shared vs.\ Independent Q/K Projection.}
We test whether decoupling should also separate query-key (Q/K) routing. In our default block, Q/K are projected from the full token, while values are projected independently from the $I$-branch and $P$-branch. The Independent-Q/K variant instead computes Q/K/V separately for the two branches, producing branch-local attention maps, but lowers performance. This suggests that shared Q/K remains useful for spatial coordination, while independent V preserves branch-specific updates (Fig.~\ref{fig:qk}).

\begin{figure}[!t]
    \centering
    \includegraphics[width=.95\linewidth]
    {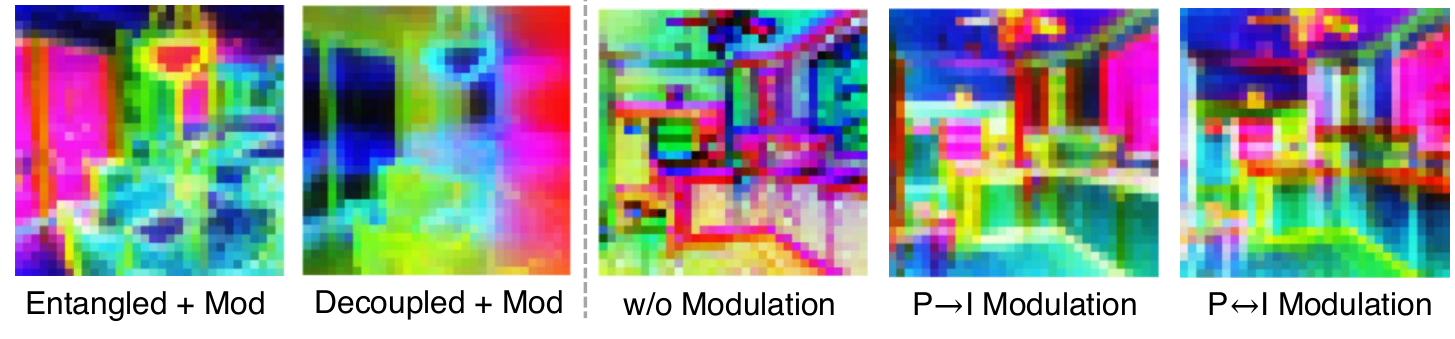}
    \vspace{-3mm}
    \caption{\textbf{Entangled vs.\ Decoupled Modulation and Bidirectional Modulation.} Left: Modulation improves performance in the decoupled structure but degrades performance in the entangled architecture. Right: Bidirectional modulation yields more structured representations.}
    \vspace{-6mm}
    \label{fig:mod}
\end{figure}

\begin{figure}[!t]
    \centering
    \begin{subfigure}[t]{0.30\linewidth}
        \centering
        \vspace{5pt}
        \includegraphics[width=\linewidth,height=30mm]{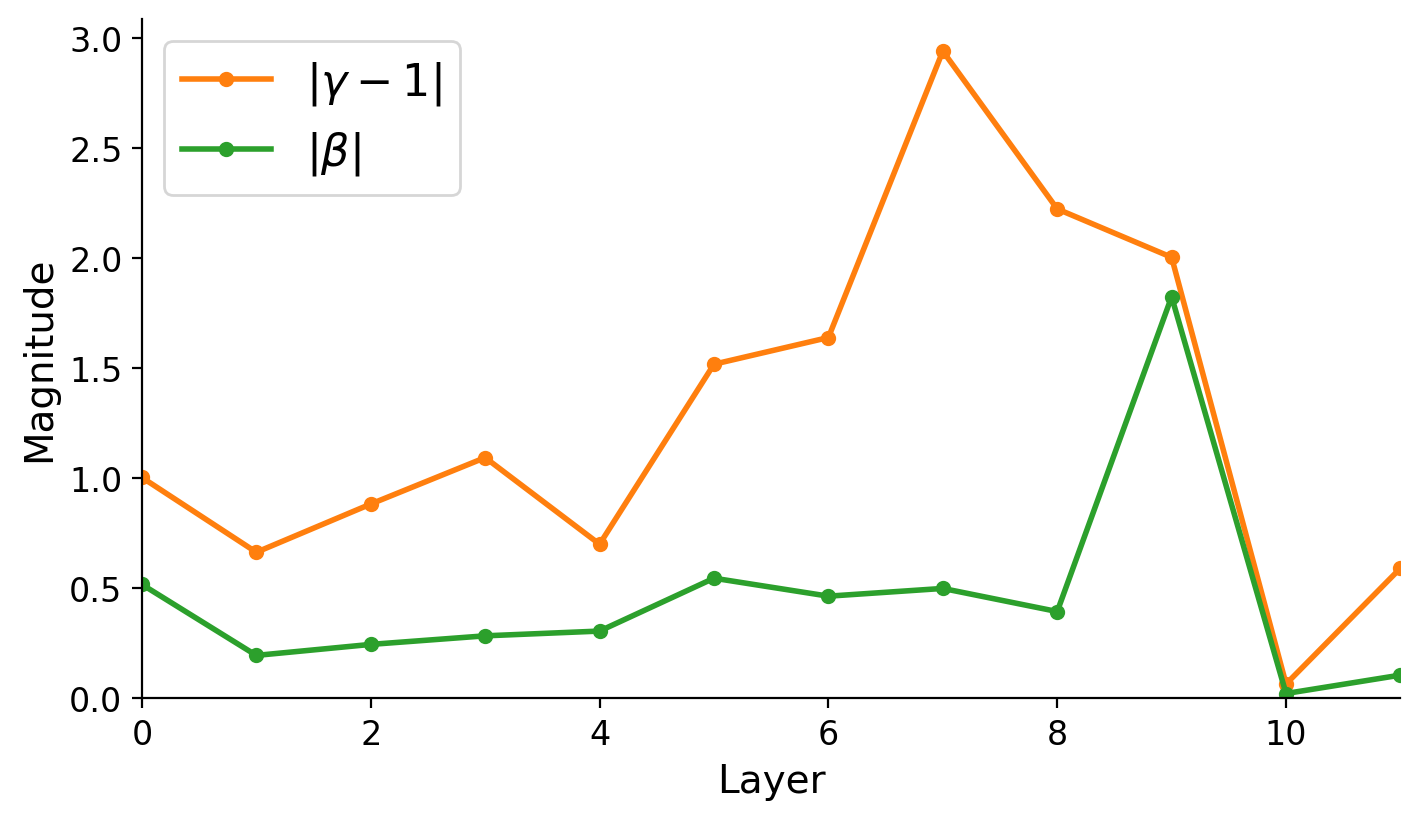}
        \caption{\textbf{Intensity of Modulation.}}
        \label{fig:film_curve}
    \end{subfigure}
    \hspace{0.012\linewidth}
    \begin{subfigure}[t]{0.30\linewidth}
        \centering
        \vspace{5pt}
        \includegraphics[width=\linewidth,height=30mm]{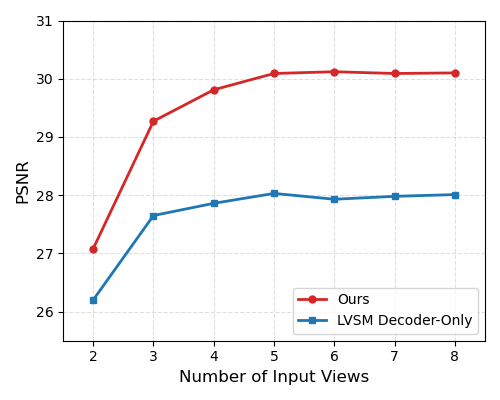}
        \caption{\textbf{Input-View Generalization.}}
        \label{fig:vary_inputs}
    \end{subfigure}
    \hspace{0.012\linewidth}
    \begin{subfigure}[t]{0.34\linewidth}
        \centering
        \vspace{5mm}
        \small
        \renewcommand{\arraystretch}{1.42}
        \setlength{\tabcolsep}{6pt}
        \begin{tabular}{l|c}
        \toprule
        \textbf{Model} & \textbf{Latency(ms)} $\downarrow$ \\
        \midrule
        Baseline & 12.2 \\
        \rowcolor{slateL}
        Decouple & 12.3 \\
        \rowcolor{slateM}
        Decouple + Mod & 13.2 \\
        \bottomrule
        \end{tabular}
        \caption{\textbf{Inference Latency.}}
        \label{fig:latency}
    \end{subfigure}
    \vspace{-6mm}
\end{figure}

\noindent\textbf{Effect of Bidirectional Modulation.}
We compare modulation in entangled and decoupled representations (Fig.~\ref{fig:mod}). In the
entangled model, affine modulation is generated from a mixed feature space, so
the scale and shift parameters cannot distinguish whether a response should
affect appearance or geometry. In the decoupled model, modulation operates between specialized branches, allowing spatial tokens to condition appearance tokens
and semantic tokens to feed back content. Fig.~\ref{fig:film_curve}
shows modulation in middle-to-late layers, with scale and shift
magnitudes peaking at blocks.

\noindent\textbf{Cross-Branch Modulation Direction.}
We further ablate the direction of cross-branch conditioning. P$\to$I modulation
uses spatial tokens to guide appearance rendering, while I$\to$P modulation uses
updated semantic tokens to refine spatial representations. P$\to$I alone is
stronger than I$\to$P alone, which matches the NVS pipeline where target-view
appearance is rendered from viewing geometry. Bidirectional modulation performs
best because the second direction provides complementary feedback after the
semantic branch has been spatially conditioned. The feature visualizations in 
Fig.~\ref{fig:mod} show that this bidirectional design yields more structured
representations.

\noindent\textbf{LayerNorm Decoupling.}
We compare branch-wise LayerNorm with Joint-LN. Joint-LN normalizes the concatenated token using shared statistics, which couples the $I$-branch and $P$-branch even after token decoupling. Branch-wise LayerNorm instead normalizes each branch with its own statistics, preserving semantic--spatial separation. As shown in Fig.~\ref{fig:scatter}, Joint-LN produces stronger cross-branch norm correlation, while Separate-LN reduces this coupling and improves reconstruction quality.

\noindent\textbf{Modulation Placement.}
We study where to insert bidirectional modulation in each block. Applying modulation before attention LayerNorm perturbs the residual stream and leads to unstable training. Applying it after the FFN is stable but weaker, because the nonlinear transformation has already been computed before cross-branch conditioning is introduced. Our final implementation starts from branch-wise normalized features and applies modulation before the branch-specific FFNs; the reverse generator receives the modulated semantic feature, as defined in Sec.~\ref{sec:film}.

\noindent\textbf{Generalization to the Number of Input Views.}The input-view generalization result in Fig.~\ref{fig:vary_inputs} shows that the design preserves its advantage when the number of test-time input views changes. This suggests that semantic-spatial decoupling improves the underlying representation rather than overfitting to a fixed context size.

\noindent\textbf{Wide-Angle Robustness.}
We evaluate multi-input trajectory spans and two-input angular separations on RealEstate10K and DL3DV~\cite{dl3dv}. Table~\ref{tab:wide_angle} shows PSNR gains in every setting, with improvements for trajectories through $180^\circ$ and a smaller gain at $270^\circ$. SSIM and LPIPS gains are less consistent under extreme viewpoints. Both models degrade with larger viewpoint changes and remain limited by target content absent from the inputs. Examples are provided in the \href{https://anonymous.4open.science/w/supplement-for-rebuttal-47B1/index.html#wide-range}{wide-angle gallery}.

\begin{table}[t]
\centering
\caption{\textbf{Wide-angle robustness.}
RE10K: RealEstate10K. $\Delta$PSNR: Ours$-$LVSM, in dB.
Bold indicates the best metric within each setting.}
\label{tab:wide_angle}

\scriptsize
\setlength{\tabcolsep}{1.2pt}

\begin{minipage}[t]{0.50\textwidth}
\vspace{0pt}
\centering
\textbf{(a) Multi-input trajectory span}\\[1mm]
\renewcommand{\arraystretch}{1.70}
\begin{tabularx}{\linewidth}{
    @{}>{\raggedright\arraybackslash}Xcclrrrr@{}
}
\toprule
\textbf{Dataset} & \textbf{Inputs} & \textbf{Span}
& \textbf{Model}
& \textbf{PSNR}$\uparrow$
& \textbf{SSIM}$\uparrow$
& \textbf{LPIPS}$\downarrow$
& $\Delta$\textbf{PSNR} \\
\midrule

\rowcolor{slateL}
RE10K & 3 & $90^\circ$ & LVSM
& 20.572 & 0.6613 & 0.2876 & --- \\
\rowcolor{slateM}
 & & & \textbf{Ours}
& \textbf{21.547} & \textbf{0.6854} & \textbf{0.2602} & +0.975 \\

\rowcolor{slateL}
RE10K & 5 & $180^\circ$ & LVSM
& 17.133 & 0.5000 & 0.4503 & --- \\
\rowcolor{slateM}
 & & & \textbf{Ours}
& \textbf{18.226} & \textbf{0.5783} & \textbf{0.3962} & +1.093 \\

\midrule

\rowcolor{slateL}
DL3DV & 3 & $90^\circ$ & LVSM
& 21.059 & 0.6849 & 0.2706 & --- \\
\rowcolor{slateM}
 & & & \textbf{Ours}
& \textbf{22.107} & \textbf{0.7534} & \textbf{0.2314} & +1.048 \\

\rowcolor{slateL}
DL3DV & 5 & $180^\circ$ & LVSM
& 16.669 & 0.5345 & 0.4665 & --- \\
\rowcolor{slateM}
 & & & \textbf{Ours}
& \textbf{18.660} & \textbf{0.6037} & \textbf{0.4055} & +1.991 \\

\rowcolor{slateL}
DL3DV & 7 & $270^\circ$ & LVSM
& 16.019 & 0.4188 & \textbf{0.5096} & --- \\
\rowcolor{slateM}
 & & & \textbf{Ours}
& \textbf{16.227} & \textbf{0.4568} & 0.5249 & +0.208 \\

\bottomrule
\end{tabularx}
\end{minipage}%
\hspace{0.015\textwidth}%
\begin{minipage}[t]{0.485\textwidth}
\vspace{0pt}
\centering
\textbf{(b) Two-input angular separation}\\[1mm]
\renewcommand{\arraystretch}{1.10}
\setlength{\tabcolsep}{1.6pt}
\begin{tabularx}{\linewidth}{
    @{}lcl*{4}{>{\raggedleft\arraybackslash}X}@{}
}
\toprule
\textbf{Dataset} & \textbf{Angle}
& \textbf{Model}
& \textbf{PSNR}$\uparrow$
& \textbf{SSIM}$\uparrow$
& \textbf{LPIPS}$\downarrow$
& $\Delta$\textbf{PSNR} \\
\midrule

\rowcolor{slateL}
RE10K & $30^\circ$ & LVSM
& 24.511 & 0.8071 & 0.1643 & --- \\
\rowcolor{slateM}
 & & \textbf{Ours}
& \textbf{25.214} & \textbf{0.8309} & \textbf{0.1465} & +0.703 \\

\rowcolor{slateL}
RE10K & $45^\circ$ & LVSM
& 22.122 & 0.7302 & 0.2291 & --- \\
\rowcolor{slateM}
 & & \textbf{Ours}
& \textbf{22.616} & \textbf{0.7503} & \textbf{0.1986} & +0.494 \\

\rowcolor{slateL}
RE10K & $90^\circ$ & LVSM
& 16.221 & 0.5522 & 0.5153 & --- \\
\rowcolor{slateM}
 & & \textbf{Ours}
& \textbf{16.792} & \textbf{0.5596} & \textbf{0.4774} & +0.571 \\

\rowcolor{slateL}
RE10K & $180^\circ$ & LVSM
& 13.304 & 0.4917 & \textbf{0.6727} & --- \\
\rowcolor{slateM}
 & & \textbf{Ours}
& \textbf{13.616} & \textbf{0.4961} & 0.7067 & +0.312 \\

\midrule

\rowcolor{slateL}
DL3DV & $30^\circ$ & LVSM
& 24.800 & 0.8128 & 0.1893 & --- \\
\rowcolor{slateM}
 & & \textbf{Ours}
& \textbf{25.352} & \textbf{0.8469} & \textbf{0.1660} & +0.552 \\

\rowcolor{slateL}
DL3DV & $45^\circ$ & LVSM
& 22.848 & 0.7719 & 0.2363 & --- \\
\rowcolor{slateM}
 & & \textbf{Ours}
& \textbf{23.315} & \textbf{0.7881} & \textbf{0.2256} & +0.467 \\

\rowcolor{slateL}
DL3DV & $90^\circ$ & LVSM
& 17.013 & \textbf{0.5982} & 0.4908 & --- \\
\rowcolor{slateM}
 & & \textbf{Ours}
& \textbf{17.036} & 0.5895 & \textbf{0.4818} & +0.023 \\

\rowcolor{slateL}
DL3DV & $180^\circ$ & LVSM
& 13.437 & 0.5308 & \textbf{0.6415} & --- \\
\rowcolor{slateM}
 & & \textbf{Ours}
& \textbf{13.848} & \textbf{0.5350} & 0.6864 & +0.411 \\

\bottomrule
\end{tabularx}
\end{minipage}
\vspace{-6mm}
\end{table}

\subsection{Computational Overhead}
\label{sec:computational_overhead}

We evaluate inference latency under the same setting for all variants, measuring
only the model forward pass after warm-up and reporting average per-sample
latency. Training-only components, including teacher models, point maps, and
auxiliary projection heads, are removed before timing.

As shown in Fig.~\ref{fig:latency}, base decoupling has nearly the same inference
latency as the baseline, because it preserves full-token Q/K attention and does
not keep supervision modules at inference. The full model adds only the
bidirectional modulation modules, resulting in a modest latency increase.
Detailed parameter derivation and timing protocol are provided in
Appendix~\ref{sec:appendix_complexity}.

\section{Conclusion}
We revisited feedforward Transformer-based novel view synthesis and identified that entangling semantic appearance and spatial conditioning within a shared token space induces representational ambiguity and grid-like distortions, which in turn degrade rendering fidelity. To address this, we propose to decouple semantic-spatial features with Independent-V attention that preserves branch purity while retaining controlled cross-stream interaction via shared routing. Extensive experiments and analysis show that the decoupling is effective for feedforward NVS Transformers.

\newpage

\bibliographystyle{unsrtnat}
\bibliography{main}

\clearpage
\appendix

\section{Implementation and Training Details}
\label{sec:appendix_impl}

\noindent\textbf{Controlled Training Budget.}
The LVSM baselines are controlled reimplementations rather than numbers
copied from the original backbone papers. This setting evaluates the relative
effect of semantic--spatial decoupling under a fixed budget; absolute baseline
numbers may differ from original reports that use different training scales,
model variants, or tuning.

\noindent\textbf{Optimizer and Learning Rate Schedule.}
Our LVSM models are trained with AdamW~\cite{loshchilov2019decoupled} ($\beta_1{=}0.9$,
$\beta_2{=}0.95$, weight decay $5\times10^{-2}$).
For scene-level training, we use a constant learning rate of $4\times10^{-4}$ with a
warm-up of 3\,000 iterations.
For object-level training, we use the same learning rate with a warm-up of 2\,500 iterations.
These LVSM experiments use mixed precision (BF16).

\noindent\textbf{Dataset-Specific Schedules.}
For scene-level training on RealEstate10K~\cite{zhou2018stereo}, we use 2 input views and 6 target views,
training on 4 A100 80G GPUs with a per-GPU batch size of 4 for 50\,K iterations
at a resolution of $256{\times}256$.
For object-level training on Objaverse~\cite{deitke2023objaverse}, we use 4 input views and 8 target views,
training on 8 A100 80G GPUs with a per-GPU batch size of 4 for 50\,K iterations.
LVSM baselines and decoupled variants use the same schedule within each benchmark.

\noindent\textbf{External Baseline Protocols.}
For deterministic sparse-view NVS on RealEstate10K, all trainable comparison
methods use the same data split, view sampling, $256{\times}256$ resolution,
four A100 GPUs, batch size, 50\,K training iterations, and fixed evaluation
views. We retain each method's native losses, optimizers, and pretrained
components. Thus, the comparison matches data, hardware, training budget,
and evaluation conditions while preserving method-specific training designs.
The methods' pretrained priors are not equated by this protocol.

\noindent\textbf{Model-Specific Configurations.}
LagerNVS~\cite{lagernvs} provides a foundation-model-assisted reference.
We follow its original design, replacing the VGGT-1B encoder with the
DA3-S DINOv2 backbone and reducing the renderer from 12 to 4 layers
to match the parameter scale. MVSplat~\cite{mvsplat} uses camera intrinsics and poses under the same RealEstate10K training and evaluation protocol.
PRoPE~\cite{prope} and raw Raymap are implemented as controlled LVSM
variants. DepthSplat~\cite{depthsplat} and pixelSplat~\cite{pixelsplat} retain their
official architectures and pretrained encoders.

\noindent\textbf{Object-Level Comparison.}
On Objaverse, LGM~\cite{lgm} and the LVSM-based models use the same training
and evaluation protocol described above.
This comparison tests transfer to object-centric scenes, where decoupling
and the optional supervision and modulation also improve reconstruction.

\noindent\textbf{Compute Resources.}
\begin{table}[h]
\centering
\vspace{-5mm}
\caption{\textbf{Training Compute.} GPU-hours are computed as wall-clock hours multiplied by the number of A100 GPUs.}
\label{tab:compute}
\small
\begin{tabular}{lccc}
\toprule
\textbf{Experiment} & \textbf{GPUs} & \textbf{Wall-clock time} & \textbf{GPU-hours} \\
\midrule
RealEstate10K Decoder-Only & 4 A100 & 8 h  & 32 \\
Objaverse Decoder-Only     & 8 A100 & 15 h & 120 \\
RealEstate10K Encoder-Decoder & 4 A100 & 12 h & 48 \\
RealEstate10K iLRM         & 4 A100 & 8 h  & 32 \\
\bottomrule
\vspace{-8mm}
\end{tabular}
\end{table}

\noindent\textbf{Loss Weights and Reconstruction Loss.}
The RGB reconstruction loss $\mathcal{L}_{\text{RGB}}$ is a weighted combination of an L2 loss and a VGG perceptual loss~\cite{johnson2016perceptual}:
\begin{equation}\setlength{\abovedisplayskip}{2pt} \setlength{\belowdisplayskip}{2pt} \setlength{\abovedisplayshortskip}{2pt} \setlength{\belowdisplayshortskip}{2pt}
  \mathcal{L}_{\text{RGB}}
  =
  \mathcal{L}_{\text{L2}}
  + \lambda_{\text{VGG}}\mathcal{L}_{\text{VGG}} .
\end{equation}
We set $\lambda_{\text{VGG}}{=}0.5$ for scene-level training and
$\lambda_{\text{VGG}}{=}1.0$ for object-level training. LPIPS is disabled in the
training objective. For supervised variants, the categorized supervision weights
are set to $\lambda_I{=}0.5$ for iREPA supervision and $\lambda_P{=}0.5$ for
geometric consistency.

\noindent\textbf{iREPA Supervision Details.}
Following iREPA~\cite{singh2025iREPA}, we use DINOv3 ViT-B/16~\cite{simeoni2025dinov3} as the frozen teacher and take its layer-8 output as the alignment target. The student feature is captured from the semantic branch at layer $k_I$. Before matching, we apply spatial normalization with $\gamma{=}0.60$ and $\epsilon{=}10^{-6}$, followed by a lightweight $3{\times}3$ convolutional projector that maps the $D/2$-dimensional semantic feature to the DINOv3 feature dimension. The matching loss is Smooth L1. The projector and DINOv3 teacher are discarded at inference.

\noindent\textbf{Geometric Grounding Details.}
For each scene or object, training views are processed offline by DA3~\cite{lin2025depthanything3} to obtain depth maps and camera parameters. We align the DA3 camera coordinates to the training camera coordinates with Umeyama registration~\cite{umeyama1991least}, back-project the depth maps into 3D point maps, and cache the resulting point maps and validity masks. During training, valid source-view 3D points are projected into target views with the aligned camera parameters, and z-buffering keeps the nearest valid correspondence for each target location. We capture spatial-branch features at layer $k_P$ and apply cosine consistency between corresponding source and target features. View pairs with fewer than 100 valid correspondences are excluded, with the threshold relaxed to 50 for Objaverse. The point maps are used only for training supervision and are not required at inference.

\section{Additional Ablations and Scaling}
\label{sec:appendix_extended_ablation}

\subsection{Component Ablations across Datasets and Architectures}
\label{sec:appendix_components}

We extend the component study to RealEstate10K and Objaverse using LVSM Decoder-Only and Encoder--Decoder architectures. The variants share the training protocol and differ only in the optional
categorized supervision and bidirectional modulation.
Table~\ref{tab:components_extended} shows that either addition improves all
three metrics over decoupling alone, and their combination performs best.
The same trend across the four settings supports the use of these components
on top of the decoupled representation.

\begin{table}[!htbp]
    \vspace{-5mm}
    \centering
    \caption{\textbf{Component ablations across datasets and architectures.}
    All configurations use decoupling; +Sup/+Mod adds categorized supervision/
    bidirectional modulation, and Full adds both.
    Bold/italics indicate the best/second-best metric within each architecture.}
    \label{tab:components_extended}
    \footnotesize
    \setlength{\tabcolsep}{2.5pt}
    \renewcommand{\arraystretch}{1.12}
    \noindent\makebox[\textwidth][c]{\ignorespaces%
    \begin{minipage}[t]{0.49\textwidth}
        \vspace{0pt}
        \centering
        \textbf{(a) RealEstate10K}\\[1mm]
        \begin{tabularx}{\linewidth}{@{}>{\raggedright\arraybackslash}X*{3}{>{\raggedleft\arraybackslash}X}@{}}
            \toprule
            \textbf{Config.} & \textbf{PSNR}$\uparrow$ & \textbf{SSIM}$\uparrow$ & \textbf{LPIPS}$\downarrow$ \\
            \midrule
            \multicolumn{4}{@{}l@{}}{\textbf{Decoder-Only}} \\
            \rowcolor{slateL}
            Decouple & 26.70 & 0.851 & 0.138 \\
            + Sup & 26.91 & 0.857 & 0.134 \\
            \rowcolor{slateM}
            + Mod & \emph{27.06} & \emph{0.860} & \emph{0.131} \\
            \rowcolor{slateH}
            \textbf{Full} & \textbf{27.21} & \textbf{0.869} & \textbf{0.125} \\
            \midrule
            \multicolumn{4}{@{}l@{}}{\textbf{Encoder--Decoder}} \\
            \rowcolor{slateL}
            Decouple & 24.84 & 0.789 & 0.170 \\
            + Sup & 25.03 & 0.795 & 0.165 \\
            \rowcolor{slateM}
            + Mod & \emph{25.17} & \emph{0.798} & \emph{0.161} \\
            \rowcolor{slateH}
            \textbf{Full} & \textbf{25.31} & \textbf{0.806} & \textbf{0.154} \\
            \bottomrule
        \end{tabularx}
    \end{minipage}%
    \unskip\hspace{0.02\textwidth}\ignorespaces%
    \begin{minipage}[t]{0.49\textwidth}
        \vspace{0pt}
        \centering
        \textbf{(b) Objaverse}\\[1mm]
        \begin{tabularx}{\linewidth}{@{}>{\raggedright\arraybackslash}X*{3}{>{\raggedleft\arraybackslash}X}@{}}
            \toprule
            \textbf{Config.} & \textbf{PSNR}$\uparrow$ & \textbf{SSIM}$\uparrow$ & \textbf{LPIPS}$\downarrow$ \\
            \midrule
            \multicolumn{4}{@{}l@{}}{\textbf{Decoder-Only}} \\
            \rowcolor{slateL}
            Decouple & 25.96 & 0.880 & 0.112 \\
            + Sup & 26.17 & 0.887 & 0.108 \\
            \rowcolor{slateM}
            + Mod & \emph{26.31} & \emph{0.890} & \emph{0.106} \\
            \rowcolor{slateH}
            \textbf{Full} & \textbf{26.46} & \textbf{0.899} & \textbf{0.101} \\
            \midrule
            \multicolumn{4}{@{}l@{}}{\textbf{Encoder--Decoder}} \\
            \rowcolor{slateL}
            Decouple & 24.16 & 0.816 & 0.137 \\
            + Sup & 24.34 & 0.822 & 0.133 \\
            \rowcolor{slateM}
            + Mod & \emph{24.48} & \emph{0.826} & \emph{0.130} \\
            \rowcolor{slateH}
            \textbf{Full} & \textbf{24.61} & \textbf{0.834} & \textbf{0.124} \\
            \bottomrule
        \end{tabularx}
    \end{minipage}%
    \unskip}
    \vspace{-3mm}
\end{table}

\subsection{Cross-View Geometry Evaluation}
\label{sec:appendix_geometry}

We evaluate whether geometric supervision improves spatial correspondence
using a foreground-depth probe on Objaverse. We freeze the decouple-only and
Decouple + Geo models and extract their layer-10 $P$-branch features.
For each foreground target patch, cosine similarity selects the most similar
source patch. We back-project the selected source patch using its mesh
ground-truth depth, transform the resulting point into the target camera
frame, and use its target-frame depth as the prediction.
This probe measures the quality of feature correspondences using source
ground-truth geometry; it does not introduce a learned depth-prediction head.

\begin{table}[!htbp]
    \vspace{-5mm}
    \centering
    \caption{\textbf{Cross-view foreground-depth evaluation on Objaverse.}
    Geo denotes spatial-branch geometric supervision.
    Bold indicates the better result for each metric.}
    \label{tab:geometry_probe}
    \footnotesize
    \setlength{\tabcolsep}{8pt}
    \renewcommand{\arraystretch}{1.15}
    \begin{tabular}{@{}lrrr@{}}
        \toprule
        \textbf{Model} & \textbf{AbsRel}$\downarrow$ & \textbf{Norm.\ RMSE}$\downarrow$ & $\boldsymbol{\delta}_1\uparrow$ \\
        \midrule
        \rowcolor{slateL}
        Decouple-only & 0.1210 & 0.0968 & 0.8535 \\
        \rowcolor{slateM}
        Decouple + Geo & \textbf{0.0988} & \textbf{0.0829} & \textbf{0.8970} \\
        \bottomrule
    \end{tabular}
    \vspace{-3mm}
\end{table}

Geometric supervision improves all three foreground-depth metrics
(Table~\ref{tab:geometry_probe}). The improvement supports more accurate
cross-view correspondences in the supervised spatial representation,
providing geometric evidence beyond the rendering metrics.

\subsection{Supervision Layer Placement}
\label{sec:appendix_sup_layer}

\begin{table}[h]
\centering
\vspace{-4mm}
\caption{\textbf{Ablation on Supervision Layer Placement.}
We vary the layer of semantic iREPA alignment and geometric grounding.}
\label{tab:sup_layer}

\scriptsize
\renewcommand{\arraystretch}{1.25}
\setlength{\tabcolsep}{5pt}

\begin{minipage}[t]{0.48\linewidth}
\centering
\textbf{(a) iREPA layer $k_I$} \\
\vspace{1mm}
\begin{tabular}{c|ccc}
\toprule
$k_I$ & PSNR$\uparrow$ & SSIM$\uparrow$ & LPIPS$\downarrow$ \\
\midrule
2  & 26.51 & 0.844 & 0.138 \\
4  & \cellcolor{slateM}26.91 & \cellcolor{slateM}0.857 & \cellcolor{slateM}0.134 \\
8  & 26.12 & 0.840 & 0.144 \\
\bottomrule
\end{tabular}\\
\vspace{1mm}
{\footnotesize Geometric supervision fixed at $k_P{=}10$.}
\end{minipage}
\hfill
\begin{minipage}[t]{0.48\linewidth}
\centering
\textbf{(b) Geometric layer $k_P$} \\
\vspace{1mm}
\begin{tabular}{c|ccc}
\toprule
$k_P$ & PSNR$\uparrow$ & SSIM$\uparrow$ & LPIPS$\downarrow$ \\
\midrule
6  & 26.61 & 0.843 & 0.138 \\
10 & \cellcolor{slateM}26.91 & \cellcolor{slateM}0.857 & \cellcolor{slateM}0.134 \\
12 & 26.47 & 0.841 & 0.140 \\
\bottomrule
\end{tabular}\\
\vspace{1mm}
{\footnotesize iREPA fixed at $k_I{=}4$.}
\end{minipage}
\vspace{-3mm}
\end{table}

This ablation is performed on the LVSM Decoder-Only model on RealEstate10K.
We study the attachment layer of the two optional branch-specific supervision
terms introduced in Sec.~\ref{sec:supervision}. When varying the semantic iREPA
layer $k_I \in \{2,4,8\}$, we keep the geometric supervision layer fixed at
$k_P{=}10$. When varying the geometric layer $k_P \in \{6,10,12\}$, we keep
the semantic iREPA layer fixed at $k_I{=}4$.

The best setting is $k_I{=}4$ and $k_P{=}10$. Very early supervision can
over-constrain low-level features before the two branches have formed stable
representations, while very late supervision is closer to the rendering head
and provides weaker guidance to intermediate representations.

\subsection{Resolution and Backbone Scaling}
\label{sec:appendix_scaling}

We test resolution scaling with LVSM Decoder-Only on RealEstate10K.
Ours denotes semantic--spatial decoupling with
bidirectional modulation, without categorized supervision.
The $512{\times}512$ models are fine-tuned from their corresponding
$256{\times}256$ checkpoints for 10K iterations with the same training
settings. In Table~\ref{tab:resolution_scaling}, Steps denotes training
iterations at resolution 256 and fine-tuning iterations at resolution 512.
Our decoupled variant improves all three rendering metrics at both
resolutions and across the reported checkpoints. The retained advantage
at resolution 512 suggests that the representation benefit persists as
more detailed image evidence becomes available.

\begin{table}[!htbp]
    \vspace{-5mm}
    \centering
    \caption{\textbf{Resolution scaling on RealEstate10K.}
    Res.\ denotes image side length in pixels.
    Bold highlights the best Ours result in each metric across the reported
    512-resolution checkpoints;
    reported $\Delta$PSNR is in dB.}
    \label{tab:resolution_scaling}
    \footnotesize
    \setlength{\tabcolsep}{2.5pt}
    \renewcommand{\arraystretch}{1.14}
    \begin{tabularx}{\linewidth}{@{}lc*{7}{>{\raggedleft\arraybackslash}X}@{}}
        \toprule
        \multirow{2}{*}{\textbf{Res.}} & \multirow{2}{*}{\textbf{Steps}}
        & \multicolumn{3}{c}{\textbf{LVSM}}
        & \multicolumn{3}{c}{\textbf{Ours}}
        & \multirow{2}{*}{$\Delta$\textbf{PSNR}} \\
        \cmidrule(lr){3-5}\cmidrule(lr){6-8}
        & & \textbf{PSNR}$\uparrow$ & \textbf{SSIM}$\uparrow$ & \textbf{LPIPS}$\downarrow$
        & \textbf{PSNR}$\uparrow$ & \textbf{SSIM}$\uparrow$ & \textbf{LPIPS}$\downarrow$ & \\
        \midrule
        256 & 10K & 23.54 & 0.7555 & 0.2205
        & 24.66 & 0.7949 & 0.1891 & +1.12 \\
        \rowcolor{slateL}
        256 & 20K & 24.86 & 0.8026 & 0.1747
        & 25.95 & 0.8360 & 0.1514 & +1.09 \\
        256 & 30K & 25.14 & 0.8142 & 0.1611
        & 26.41 & 0.8475 & 0.1409 & +1.27 \\
        \rowcolor{slateL}
        256 & 40K & 25.22 & 0.8066 & 0.1553
        & 26.93 & 0.8582 & 0.1341 & +1.71 \\
        \rowcolor{slateM}
        256 & 50K & 26.10 & 0.8387 & 0.1444
        & 27.06 & 0.8603 & 0.1313 & +0.96 \\
        \midrule
        512 & 2K & 27.20 & 0.8656 & 0.1808
        & 27.97 & 0.8795 & 0.1722 & +0.77 \\
        \rowcolor{slateL}
        512 & 4K & 27.30 & 0.8665 & 0.1786
        & 28.02 & 0.8801 & 0.1714 & +0.72 \\
        512 & 6K & 27.26 & 0.8676 & 0.1767
        & 28.03 & 0.8815 & 0.1698 & +0.77 \\
        \rowcolor{slateL}
        512 & 8K & 27.47 & 0.8707 & 0.1770
        & \textbf{28.28} & 0.8833 & 0.1682 & +0.81 \\
        \rowcolor{slateH}
        512 & 10K & 27.45 & 0.8712 & 0.1759
        & 28.19 & \textbf{0.8834} & \textbf{0.1681} & +0.74 \\
        \bottomrule
    \end{tabularx}
    \vspace{-4mm}
\end{table}

We also train a 24-layer LVSM Decoder-Only backbone under the same
RealEstate10K protocol. Ours uses decoupling with bidirectional modulation
and no categorized supervision. Table~\ref{tab:backbone_scaling} shows improvements
throughout training, indicating that additional model capacity does not
remove the benefit of decoupling in this setting. The 50K results are also
included in Table~\ref{tab:main}. These experiments test resolution and
backbone depth under the stated budgets; broader training-scale validation
remains a direction for future work.

\begin{table}[!htbp]
    \vspace{-5mm}
    \centering
    \caption{\textbf{Backbone scaling on RealEstate10K.}
    Both models use a 24-layer decoder-only backbone;
    Ours adds decoupling and modulation without supervision.
    Bold/italics indicate the best/second-best result in each metric column
    across training checkpoints.}
    \label{tab:backbone_scaling}
    \footnotesize
    \setlength{\tabcolsep}{3pt}
    \renewcommand{\arraystretch}{1.14}
    \begin{tabularx}{\linewidth}{@{}c*{7}{>{\raggedleft\arraybackslash}X}@{}}
        \toprule
        \multirow{2}{*}{\textbf{Steps}}
        & \multicolumn{3}{c}{\textbf{LVSM}}
        & \multicolumn{3}{c}{\textbf{Ours}}
        & \multirow{2}{*}{$\Delta$\textbf{PSNR}} \\
        \cmidrule(lr){2-4}\cmidrule(lr){5-7}
        & \textbf{PSNR}$\uparrow$ & \textbf{SSIM}$\uparrow$ & \textbf{LPIPS}$\downarrow$
        & \textbf{PSNR}$\uparrow$ & \textbf{SSIM}$\uparrow$ & \textbf{LPIPS}$\downarrow$ & \\
        \midrule
        10K & 23.72 & 0.7637 & 0.2143
        & 25.27 & 0.8152 & 0.1729 & +1.55 \\
        \rowcolor{slateL}
        20K & 25.04 & 0.8044 & 0.1715
        & 26.36 & 0.8452 & 0.1426 & +1.32 \\
        30K & 25.47 & 0.8213 & 0.1557
        & 26.66 & 0.8537 & 0.1326 & +1.19 \\
        \rowcolor{slateM}
        40K & \emph{25.76} & \emph{0.8264} & \emph{0.1497}
        & \emph{26.98} & \emph{0.8598} & \emph{0.1304} & +1.22 \\
        \rowcolor{slateH}
        50K & \textbf{26.18} & \textbf{0.8420} & \textbf{0.1406}
        & \textbf{27.37} & \textbf{0.8671} & \textbf{0.1249} & +1.19 \\
        \bottomrule
    \end{tabularx}
    \vspace{-5mm}
\end{table}

\subsection{Generalization to the iLRM Architecture}
\label{sec:appendix_ilrm}

We further apply decoupling with bidirectional modulation to iLRM~\cite{kang2025ilrmiterativelarge3d},
a feed-forward 3D reconstruction model that predicts 3D Gaussian Splatting
parameters. Unlike LVSM, iLRM refines image-free Gaussian query tokens through
alternating Read blocks and Self-Attention blocks. Read blocks let query tokens
attend to context-view support tokens, while Self-Attention blocks propagate
interactions among the query tokens.

In our controlled iLRM setting, the processor uses 12 alternating blocks with
patch size 8, token dimension 768, and head dimension 64. We decouple both the
query stream and the support stream. Query tokens use zero-initialized semantic
branches and Pl\"ucker-initialized spatial branches; support tokens use RGB
patches for the semantic branch and Pl\"ucker-ray patches for the spatial
branch. The Read and Self-Attention blocks then use the same Independent-V,
branch-wise normalization, branch-specific FFN, and bidirectional modulation
design described in Sec.~\ref{sec:decouple} and Sec.~\ref{sec:film}. The final token is
decoded by the original iLRM Gaussian head, leaving the output representation
unchanged.

We train the iLRM baseline and decoupled variant on RealEstate10K under the same
controlled protocol. We use 2 context views and 4 target views at
$256{\times}256$ resolution. Both models use AdamW, learning rate
$4\times10^{-4}$, 2\,500 warm-up iterations, gradient clipping at 1.0, BF16
mixed precision, 4 A100 80G GPUs, per-GPU batch size 32, and 5\,K training
iterations. The reconstruction objective follows the iLRM configuration and uses
RGB MSE loss. No teacher model, point map, or auxiliary projection head is used
at inference.

Under this matched setting, Table~\ref{tab:ilrm_comp} shows improvements in all
three metrics, supporting transfer of the combined decoupling and modulation
design to iLRM.

\begin{table}[h]
\centering
\vspace{-5mm}
\caption{\textbf{Quantitative Comparison on iLRM.} Decoupling with bidirectional modulation improves reconstruction under the matched training protocol.}
\label{tab:ilrm_comp}
\begin{tabular}{l|ccc}
\toprule
\textbf{Model} & \textbf{PSNR} $\uparrow$ & \textbf{SSIM} $\uparrow$ & \textbf{LPIPS} $\downarrow$ \\
\midrule
iLRM baseline & 22.42 & 0.764 & 0.276 \\
\textbf{iLRM-Decouple + Mod (Ours)} & \cellcolor{slateM}23.15 & \cellcolor{slateM}0.789 & \cellcolor{slateM}0.245 \\
\bottomrule
\end{tabular}
\vspace{-4mm}
\end{table}

\subsection{Complementary Generative Comparisons}
\label{sec:appendix_generative}

We compare with adapted GEN3C~\cite{gen3c} and
Lyra~\cite{lyra}. Their input protocols, pretrained priors, and model scales
differ from those of feedforward reconstruction.
We therefore report these experiments as complementary references and
separate adapted models from published full-scale results.

\noindent\textbf{GEN3C.}
We replace its large video-diffusion backbone with MobileI2V, retaining
the independent 3D-cache construction and in-layer cache fusion.
We adapt the latent projections and temporal/spatial interfaces to the new
backbone and train this variant on the same RealEstate10K split for
50\,K iterations.
Table~\ref{tab:generative_reference}(a) shows that our model improves all
three metrics over this adapted variant with fewer parameters.
The published full-scale GEN3C result has lower LPIPS under its original
protocol and is included as a contextual reference.

\noindent\textbf{Lyra.}
We follow its static single-image-to-3D setting: one source image is
provided, while the model internally generates and fuses multiple
camera-controlled trajectories.
Our GEN3C-MobileI2V model serves as a frozen teacher, and we train Lyra's
3DGS decoder for 50\,K iterations using trajectories generated from
RealEstate10K source images.
Pl\"ucker camera conditioning, multi-trajectory fusion, and static 3DGS
reconstruction are retained.
Table~\ref{tab:generative_reference}(b) shows that our model improves over
the adapted Lyra model, while the published full-scale Lyra result is
stronger under its original setting.
The latter uses a roughly 7B-parameter pretrained video model, longer and
higher-resolution generated trajectories, and broader pretraining.

\begin{table}[!htbp]
    \vspace{-5mm}
    \centering
    \caption{\textbf{Complementary generative comparisons on RealEstate10K.}
    Adapted denotes MobileI2V-based variants; Params are in millions.
    $^{*}$ denotes published results under the original protocols.
    Bold compares Ours with the adapted model only.}
    \label{tab:generative_reference}
    \footnotesize
    \setlength{\tabcolsep}{2pt}
    \renewcommand{\arraystretch}{1.15}
    \noindent\makebox[\textwidth][c]{\ignorespaces%
    \begin{minipage}[t]{0.49\textwidth}
        \vspace{0pt}
        \centering
        \textbf{(a) GEN3C}\\[1mm]
        \begin{tabularx}{\linewidth}{@{}>{\raggedright\arraybackslash}Xrrrr@{}}
            \toprule
            \textbf{Model} & \textbf{Params} & \textbf{PSNR}$\uparrow$ & \textbf{SSIM}$\uparrow$ & \textbf{LPIPS}$\downarrow$ \\
            \midrule
            Adapted & 267.4 & 20.57 & 0.652 & 0.318 \\
            \rowcolor{slateL}
            Published$^{*}$ & 1521 & 24.08 & 0.860 & 0.110 \\
            \rowcolor{slateM}
            \textbf{Ours} & 57.2 & \textbf{27.21} & \textbf{0.869} & \textbf{0.125} \\
            \bottomrule
        \end{tabularx}
    \end{minipage}%
    \unskip\hspace{0.02\textwidth}\ignorespaces%
    \begin{minipage}[t]{0.49\textwidth}
        \vspace{0pt}
        \centering
        \textbf{(b) Lyra (single-image input)}\\[1mm]
        \begin{tabularx}{\linewidth}{@{}>{\raggedright\arraybackslash}Xrrrr@{}}
            \toprule
            \textbf{Model} & \textbf{Params} & \textbf{PSNR}$\uparrow$ & \textbf{SSIM}$\uparrow$ & \textbf{LPIPS}$\downarrow$ \\
            \midrule
            Adapted & 304.6 & 17.71 & 0.570 & 0.364 \\
            \rowcolor{slateL}
            Published$^{*}$ & 7033 & 21.79 & 0.752 & 0.219 \\
            \rowcolor{slateM}
            \textbf{Ours} & 57.2 & \textbf{18.65} & \textbf{0.589} & \textbf{0.328} \\
            \bottomrule
        \end{tabularx}
    \end{minipage}%
    \unskip}
    \vspace{-4mm}
\end{table}

These comparisons clarify the scope of our contribution: decoupling improves
observation-faithful reconstruction without iterative diffusion sampling,
whereas generative priors can support perceptual quality and plausible
completion of unobserved content.
The two directions can also be combined, as in pipelines that use generated
views for reconstruction~\cite{lgm,cat3d,lyra}.
Related generative approaches, including Lyra 2.0~\cite{lyra2},
TRELLIS~\cite{trellis}, and Hunyuan3D 2.1~\cite{hunyuan3d21}, also use
different inputs, outputs, and evaluation objectives.
These differences call for task-specific evaluation when relating their
progress to feedforward NVS.

\section{Representation Diagnostics and Visualization}
\label{sec:appendix_artifacts}

\subsection{Quantifying Grid-Like Artifacts}

We define the axis-aligned Grid Artifact Score (GAS) as
\begin{equation}
    \mathrm{GAS}=10\log_{10}\left(\frac{E_{\mathrm{axis}}}{E_{\mathrm{diag}}}\right),
\end{equation}
where $E_{\mathrm{axis}}$ is the mid/high-frequency spectral energy near
the horizontal and vertical axes, and $E_{\mathrm{diag}}$ is the energy
in matched diagonal sectors.
Higher GAS indicates stronger axis-aligned grid structure.
On 128 synthetic $1/f$ feature backgrounds, GAS increases monotonically
with injected grid strength.
Adding white noise or low-frequency structure changes GAS by less than
0.12\,dB and 0.01\,dB, respectively.

Table~\ref{tab:gas_association} combines the layer-wise measurements and
the rendering association study on RealEstate10K.
Register tokens do not reduce GAS in the entangled representation.
Decoupling reduces it in the semantic $I$-branch, while the spatial
$P$-branch retains strong axis-aligned structure.
This distinction supports separating spatial structure from appearance
features, rather than suppressing it in both branches.
Across the five rendering variants, final-layer semantic GAS has Spearman
correlation $\rho=-0.700$ with PSNR and $\rho=+0.700$ with LPIPS.
This association concerns these five variants and alone does not establish
a causal relationship.

\begin{table}[!htbp]
    \vspace{-5mm}
    \centering
    \footnotesize
    \renewcommand{\arraystretch}{1.15}
    \noindent\makebox[\textwidth][c]{\ignorespaces%
    \begin{minipage}[t]{0.50\textwidth}
        \vspace{0pt}
        \centering
        \caption{\textbf{Grid artifacts on RealEstate10K.}
        GAS (dB) at layers 8/12; $I/P$: semantic/spatial.}
        \label{tab:gas_association}
        \setlength{\tabcolsep}{1.75pt}
        \begin{tabularx}{\linewidth}{@{}>{\raggedright\arraybackslash}Xrrrr@{}}
            \toprule
            \textbf{Variant} & \textbf{L8} & \textbf{L12} & \textbf{PSNR}$\uparrow$ & \textbf{LPIPS}$\downarrow$ \\
            \midrule
            Baseline & 13.11 & 5.56 & 26.10 & 0.144 \\
            Baseline + Registers & 13.85 & 5.67 & 26.04 & 0.146 \\
            Decouple + Joint LN ($I$) & --- & 4.55 & 26.01 & 0.149 \\
            \rowcolor{slateM}
            Decouple ($I$) & 5.06 & \emph{4.21} & \emph{26.70} & \emph{0.138} \\
            \rowcolor{slateH}
            Decouple + Modulation ($I$) & --- & \textbf{3.88} & \textbf{27.06} & \textbf{0.131} \\
            \midrule
            \rowcolor{slateL}
            Decouple ($P$) & 19.18 & 17.05 & --- & --- \\
            \bottomrule
        \end{tabularx}
    \end{minipage}%
    \unskip\hspace{0.02\textwidth}\ignorespaces%
    \begin{minipage}[t]{0.48\textwidth}
        \vspace{0pt}
        \centering
        \caption{\textbf{Layer-8 feature intervention.}
        A/D: axial/diagonal; $\Delta$GAS and $\Delta$PSNR in dB.}
        \label{tab:artifact_intervention}
        \setlength{\tabcolsep}{1.25pt}
        \begin{tabularx}{\linewidth}{@{}>{\raggedright\arraybackslash}Xrrrrr@{}}
            \toprule
            $\boldsymbol{\alpha}$ &
            \shortstack{\textbf{A}\\$\Delta$GAS} &
            \shortstack{\textbf{A}\\$\Delta$PSNR} &
            \shortstack{\textbf{D}\\$\Delta$PSNR} &
            \shortstack{\textbf{A$-$D}\\$\Delta$PSNR} &
            \shortstack{\textbf{RMS}\\\textbf{(A/D)}} \\
            \midrule
            0.02 & +0.24 & -0.002 & -0.003 & +0.001 & 0.0165/0.0199 \\
            \rowcolor{slateL}
            0.05 & +0.59 & -0.031 & -0.022 & -0.009 & 0.0407/0.0495 \\
            \rowcolor{slateM}
            0.10 & +1.15 & -0.140 & -0.095 & -0.045 & 0.0796/0.0984 \\
            \rowcolor{slateH}
            0.20 & +2.19 & -0.507 & -0.394 & -0.113 & 0.1515/0.1933 \\
            \bottomrule
        \end{tabularx}
    \end{minipage}%
    \unskip}
    \vspace{-4mm}
\end{table}

\subsection{Controlled Feature Intervention}

We intervene at layer 8 of the entangled baseline.
At each strength $\alpha$, we inject either an axis-aligned component or
a diagonal component from the same frequency band.
The diagonal injection controls for the general effect of perturbing
features, while the axial injection increases grid-like structure.
Table~\ref{tab:artifact_intervention} reports changes relative to the
unmodified baseline and the RMS feature changes.

For $\alpha\geq0.05$, axial injection causes a larger PSNR reduction
than the diagonal control despite a smaller RMS feature change.
The excess degradation grows with intervention strength.
At $\alpha=0.02$, this difference is absent.
Together, the results provide controlled evidence that increasing
axis-aligned artifacts can harm rendering beyond the general effect of
feature perturbations, without implying that all reconstruction errors
arise from these artifacts.
Layer-wise GAS curves and intervention plots are available in the
\href{https://anonymous.4open.science/w/supplement-for-rebuttal-47B1/index.html#artifacts}{online artifact analysis}.

\subsection{Layer-Wise Feature Visualization}
\begin{figure}[h]
    \centering
    \includegraphics[width=0.95\linewidth]{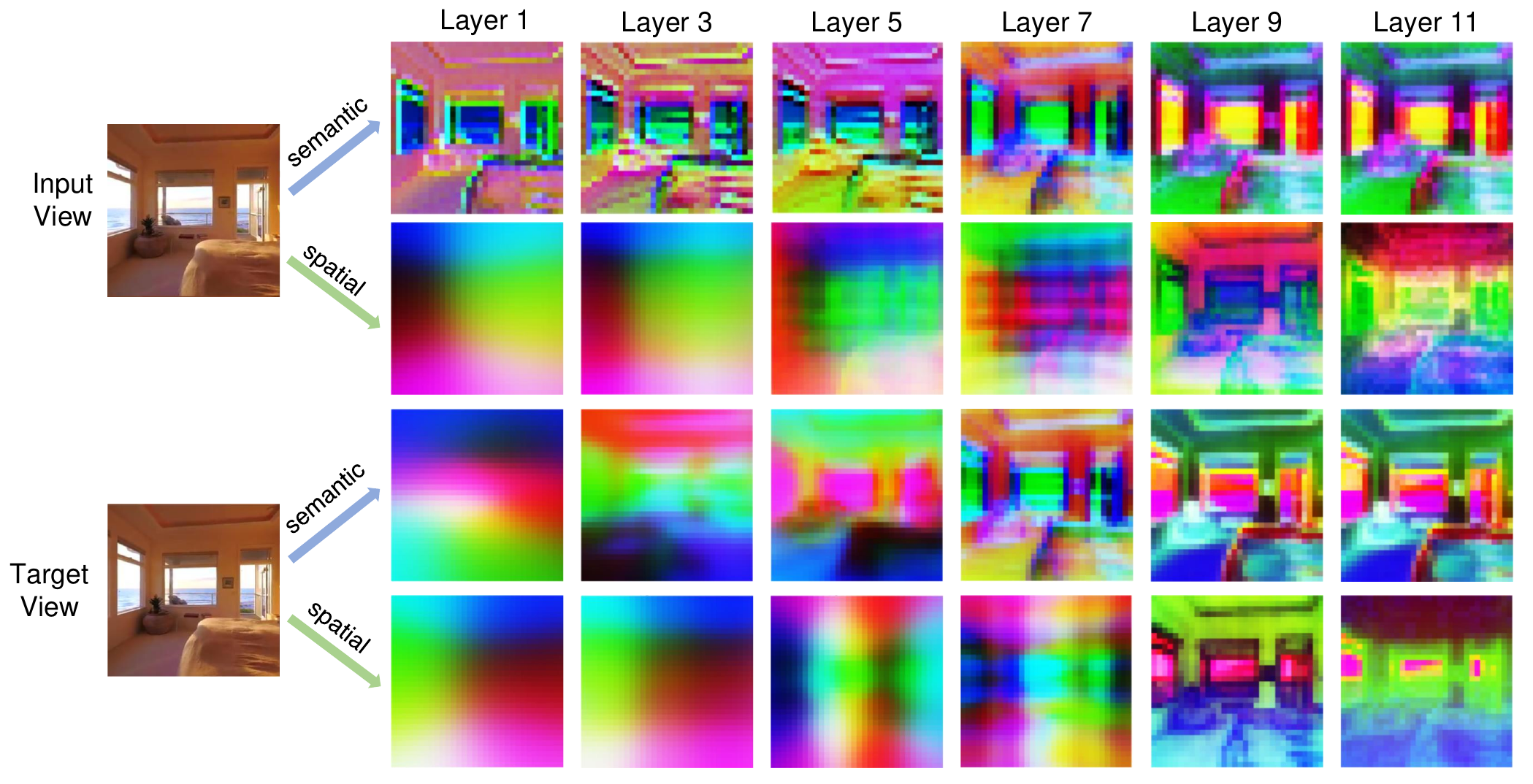}
    \caption{\textbf{Layer-wise Feature Visualization of Decouple-Only.}
    PCA visualizations of intermediate input-view and target-view features from the decouple-only model.}
    \vspace{-5mm}
    \label{fig:gen}
\end{figure}

\noindent\textbf{Layer-wise Feature Evolution.}
Fig.~\ref{fig:gen} provides PCA visualizations of input-view and target-view features from the decouple-only model across multiple Transformer layers. The visualizations show that the decoupled representation produces organized feature structures throughout the network.

\subsection{Qualitative Comparisons and Failure Cases}
\label{sec:appendix_boundary}

The \href{https://anonymous.4open.science/w/supplement-for-rebuttal-47B1/index.html#qualitative}{online qualitative comparisons}
contain 12 RealEstate10K scenes and 12 Objaverse objects, with three novel
views for each example.
They complement Fig.~\ref{fig:infer_compare} with sharper contours and
reduced blur across varied scenes and objects.
\href{https://anonymous.4open.science/w/supplement-for-rebuttal-47B1/index.html#video-results}{Rendering videos}
provide additional trajectory visualizations.

\noindent\textbf{Challenging viewpoints and visibility.}
The \href{https://anonymous.4open.science/w/supplement-for-rebuttal-47B1/index.html#wide-range}{wide-angle examples}
accompany Table~\ref{tab:wide_angle}.
Although PSNR improves in every tested setting, the advantage shrinks
under the extreme $270^\circ$ trajectory, and LPIPS worsens in this
setting and in both $180^\circ$ two-input settings.
When target content is observed but input-view overlap is low, both
models can blur details, with decoupling generally reducing the blur.
When target regions are absent from every input, both may produce gray
or meaningless pixels because the necessary visual evidence is missing.

\noindent\textbf{Complex structures.}
Fine structures, cluttered backgrounds, severe occlusions, and repetitive
textures remain challenging.
Our model generally produces sharper contours, but both models can still
generate broken structures or inaccurate textures, as shown in the
\href{https://anonymous.4open.science/w/supplement-for-rebuttal-47B1/index.html#failure-cases}{failure-case gallery}.
Decoupling improves the use of available evidence; it does not resolve
the ambiguity of unobserved content in deterministic NVS.

\section{Inference Complexity Analysis}
\label{sec:appendix_complexity}

We provide the parameter derivation for Sec.~\ref{sec:computational_overhead}. Let
$D$ denote the token dimension and $r$ the FFN expansion ratio. Bias terms are
omitted because they are lower-order $O(D)$ terms.

\noindent\textbf{Baseline Block.}
For the entangled baseline, attention uses QKV and output projections, and the
FFN uses hidden dimension $rD$:
\begin{equation}
  \mathcal{P}_{\text{attn}}^{\text{base}}=4D^2,
  \qquad
  \mathcal{P}_{\text{ffn}}^{\text{base}}=2rD^2 .
\end{equation}
Thus,
\begin{equation}
  \mathcal{P}_{\text{block}}^{\text{base}}
  =
  (2r+4)D^2 + O(D).
\end{equation}

\noindent\textbf{Base Decoupled Block.}
In the base decoupled block, Q/K are shared over the full token, while value,
output, and FFN transformations are branch-specific:
\begin{equation}
  \mathcal{P}_{\text{attn}}^{\text{dec}}=3D^2,
  \qquad
  \mathcal{P}_{\text{ffn}}^{\text{dec}}=rD^2 .
\end{equation}
Therefore,
\begin{equation}
  \mathcal{P}_{\text{block}}^{\text{dec}}
  =
  (r+3)D^2 + O(D),
\end{equation}
which is no larger than the baseline block.

\noindent\textbf{LayerNorm and Modulation.}
Branch-wise LayerNorm preserves parameter count because two full-dimensional
LayerNorms are replaced by four half-dimensional LayerNorms:
\begin{equation}
  2D = 4\cdot\frac{D}{2}.
\end{equation}
Bidirectional modulation adds two FiLM generators:
\begin{equation}
  \mathcal{P}_{\text{FiLM}}
  =
  2\left(\frac{D}{2}\cdot D + D\right)
  =
  D^2 + 2D .
\end{equation}
With $D{=}768$, this adds about $0.59$M parameters per block, or about $7.1$M
parameters across 12 layers.

\noindent\textbf{Latency Protocol.}
Latency is measured under the same evaluation setting for all variants. We
exclude warm-up batches and report average per-sample forward time. Image
saving, metric computation, teacher models, point maps, and auxiliary heads are
not included. Categorized supervision therefore does not affect inference
latency, while bidirectional modulation adds the only extra inference-time
projections.

\section{Limitations and Future Directions}
Our evaluation uses controlled feedforward NVS budgets. Unobserved target content,
extreme camera changes, and complex structures remain challenging
(Appendix~\ref{sec:appendix_boundary}). Future work includes larger training scales,
more diverse outdoor and dynamic scenes, and longer trajectories; extending the
design to other camera encodings and reconstruction backbones; and developing
lighter supervision to reduce offline preprocessing.

\section{Broader Impacts}\label{sec:impacts}
Improved feedforward reconstruction may benefit 3D content creation, AR/VR, and
robotics simulation. It could also be misused to create misleading imagery,
reconstruct private spaces, or generate unauthorized 3D representations.
We recommend public, licensed, or consented training data, permission for
privacy-sensitive applications, and clear disclosure of synthetic outputs.
For planned code and model releases, we plan to document intended research use,
dataset assumptions, and known limitations, and discourage privacy-invasive
or misleading applications.

\section{Existing Assets, Licenses, and Terms}\label{sec:assets}
We use existing assets only for research and cite their original sources.
RealEstate10K camera trajectories are released by Google LLC under CC BY 4.0;
video frames remain subject to their source terms. Objaverse uses ODC-By v1.0,
while objects retain their metadata licenses: CC BY 4.0, CC BY-NC 4.0,
CC BY-NC-SA 4.0, CC BY-SA 4.0, or CC0 1.0. We follow asset-specific terms
and do not redistribute the original objects.

For training-time supervision, DINOv3 ViT-B/16 uses the DINOv3 License;
DA3NESTED-GIANT-LARGE uses CC BY-NC 4.0, and DA3 code uses Apache-2.0.
Their checkpoints, intermediate caches, and teacher outputs are not
redistributed with this submission.

\newpage
\section*{NeurIPS Paper Checklist}

\begin{enumerate}

\item {\bf Claims}
    \item[] Question: Do the main claims made in the abstract and introduction accurately reflect the paper's contributions and scope?
    \item[] Answer: \answerYes{} 
    \item[] Justification: The abstract and introduction distinguish semantic--spatial decoupling from optional supervision and modulation, with supporting methods and controlled experiments in Sections~\ref{sec:method} and~\ref{sec:experiments}. Appendix~\ref{sec:appendix_generative} separately identifies adapted generative comparisons and published results under different protocols.

\item {\bf Limitations}
    \item[] Question: Does the paper discuss the limitations of the work performed by the authors?
    \item[] Answer: \answerYes{} 
    \item[] Justification: Section~\ref{sec:experiments} and Appendix~\ref{sec:appendix_boundary} discuss extreme viewpoints, low overlap, unobserved content, and complex structures, including settings where SSIM or LPIPS does not improve. The appendix also includes a Limitations and Future Directions section addressing the controlled training scale and remaining extensions.

\item {\bf Theory assumptions and proofs}
    \item[] Question: For each theoretical result, does the paper provide the full set of assumptions and a complete (and correct) proof?
    \item[] Answer: \answerNA{} 
    \item[] Justification: The paper does not present theoretical results, theorems, or formal proofs; its contribution is architectural and empirical.

    \item {\bf Experimental result reproducibility}
    \item[] Question: Does the paper fully disclose all the information needed to reproduce the main experimental results of the paper to the extent that it affects the main claims and/or conclusions of the paper (regardless of whether the code and data are provided or not)?
    \item[] Answer: \answerYes{} 
    \item[] Justification: Sections~\ref{sec:method} and~\ref{sec:experiments}, together with Appendix~\ref{sec:appendix_impl}, describe the architecture, objectives, supervision construction, and main benchmark protocols. Appendix~\ref{sec:appendix_extended_ablation} provides the additional ablation, scaling, iLRM, and adapted generative settings.

\item {\bf Open access to data and code}
    \item[] Question: Does the paper provide open access to the data and code, with sufficient instructions to faithfully reproduce the main experimental results, as described in supplemental material?
    \item[] Answer: \answerYes{}
    \item[] Justification: The \href{https://github.com/hangzay/ssd_lvsm}{public repository}, linked from the project page, releases the core semantic--spatial decoupling implementation, configurations, and installation, training, and evaluation instructions for the main LVSM experiments. We use publicly available RealEstate10K and Objaverse data, with benchmark protocols described in Section~\ref{sec:experiments} and Appendix~\ref{sec:appendix_impl}.

\item {\bf Experimental setting/details}
    \item[] Question: Does the paper specify all the training and test details (e.g., data splits, hyperparameters, how they were chosen, type of optimizer) necessary to understand the results?
    \item[] Answer: \answerYes{} 
    \item[] Justification: Section~\ref{sec:experiments} and Appendix~\ref{sec:appendix_impl} specify the main datasets, view sampling, resolution, optimizer, loss weights, batch sizes, training schedules, and metrics, while distinguishing external methods' native training designs. Appendix~\ref{sec:appendix_extended_ablation} gives additional protocols and the supervision-layer selection study.

\item {\bf Experiment statistical significance}
    \item[] Question: Does the paper report error bars suitably and correctly defined or other appropriate information about the statistical significance of the experiments?
    \item[] Answer: \answerNo{} 
    \item[] Justification: The reported results do not include error bars, confidence intervals, or statistical significance tests across independent training runs. Trends across metrics, architectures, and datasets are empirical support, rather than a claim of statistical significance.

\item {\bf Experiments compute resources}
    \item[] Question: For each experiment, does the paper provide sufficient information on the computer resources (type of compute workers, memory, time of execution) needed to reproduce the experiments?
    \item[] Answer: \answerYes{}
    \item[] Justification: Appendix~\ref{sec:appendix_impl} specifies GPU type, GPU memory, GPU counts, batch sizes, and training schedules; Table~\ref{tab:compute} reports wall-clock training time and GPU-hours for the principal training settings. Additional training settings are described in Appendix~\ref{sec:appendix_extended_ablation}.
    
\item {\bf Code of ethics}
    \item[] Question: Does the research conducted in the paper conform, in every respect, with the NeurIPS Code of Ethics \url{https://neurips.cc/public/EthicsGuidelines}?
    \item[] Answer: \answerYes{} 
    \item[] Justification: The study uses credited research datasets and pretrained models, and does not involve crowdsourcing or direct human-subject experiments. Appendices~\ref{sec:assets} and~\ref{sec:impacts} describe asset usage and discuss privacy-sensitive and misleading applications.

\item {\bf Broader impacts}
    \item[] Question: Does the paper discuss both potential positive societal impacts and negative societal impacts of the work performed?
    \item[] Answer: \answerYes{} 
    \item[] Justification: Appendix~\ref{sec:impacts} discusses benefits for 3D content creation, AR/VR, and robotics simulation, alongside misleading imagery, reconstruction of private spaces, and unauthorized 3D representations. It also recommends appropriate data use, permission for privacy-sensitive applications, and disclosure of synthetic outputs.
    
\item {\bf Safeguards}
    \item[] Question: Does the paper describe safeguards that have been put in place for responsible release of data or models that have a high risk for misuse (e.g., pre-trained language models, image generators, or scraped datasets)?
    \item[] Answer: \answerNo{}
    \item[] Justification: Appendix~\ref{sec:impacts} discusses potential misuse and recommends responsible data use and release documentation. It does not report implemented access controls, safety filters, or other release safeguards.

\item {\bf Licenses for existing assets}
    \item[] Question: Are the creators or original owners of assets (e.g., code, data, models), used in the paper, properly credited and are the license and terms of use explicitly mentioned and properly respected?
    \item[] Answer: \answerNo{}
    \item[] Justification: Appendix~\ref{sec:assets} names the licenses and terms for RealEstate10K, Objaverse, DINOv3, and DA3, and the paper credits the other datasets and methods. The manuscript does not explicitly enumerate licenses and versions for every additional dataset, baseline implementation, pretrained component, and evaluation asset.

\item {\bf New assets}
    \item[] Question: Are new assets introduced in the paper well documented and is the documentation provided alongside the assets?
    \item[] Answer: \answerYes{}
    \item[] Justification: The \href{https://github.com/hangzay/ssd_lvsm}{released code} includes a README describing dependencies, data preparation, supported configurations, and training and evaluation commands. Its \href{https://github.com/hangzay/ssd_lvsm/blob/main/LICENSE.md}{license file} states the project-code license and distinguishes third-party code and data terms.

\item {\bf Crowdsourcing and research with human subjects}
    \item[] Question: For crowdsourcing experiments and research with human subjects, does the paper include the full text of instructions given to participants and screenshots, if applicable, as well as details about compensation (if any)? 
    \item[] Answer: \answerNA{} 
    \item[] Justification: The work does not involve crowdsourcing experiments or research with human subjects.

\item {\bf Institutional review board (IRB) approvals or equivalent for research with human subjects}
    \item[] Question: Does the paper describe potential risks incurred by study participants, whether such risks were disclosed to the subjects, and whether Institutional Review Board (IRB) approvals (or an equivalent approval/review based on the requirements of your country or institution) were obtained?
    \item[] Answer: \answerNA{} 
    \item[] Justification: The work does not involve crowdsourcing or human-subject research, so IRB approval or equivalent review is not applicable.

\item {\bf Declaration of LLM usage}
    \item[] Question: Does the paper describe the usage of LLMs if it is an important, original, or non-standard component of the core methods in this research? Note that if the LLM is used only for writing, editing, or formatting purposes and does \emph{not} impact the core methodology, scientific rigor, or originality of the research, declaration is not required.
    \item[] Answer: \answerNA{} 
    \item[] Justification: LLM assistance is limited to writing, editing, and formatting the manuscript. LLMs are not part of the core method or experimental pipeline.

\end{enumerate}

\end{document}